\pdfoutput=1

\documentclass[11pt]{article}

\usepackage[preprint]{acl}

\usepackage{times}
\usepackage{latexsym}
\usepackage{amsmath}
\usepackage{amssymb}
\usepackage{enumitem}
\usepackage[most]{tcolorbox}
\usepackage{multirow}
\usepackage{array}
\usepackage{adjustbox}
\usepackage[T1]{fontenc}
\usepackage[utf8]{inputenc}
\usepackage{microtype}
\usepackage{inconsolata}
\usepackage{graphicx}
\usepackage{booktabs}
\usepackage{comment}
\usepackage{subcaption}

\newcommand\blfootnote[1]{%
  \begingroup
  \renewcommand\thefootnote{}\footnote{#1}%
  \addtocounter{footnote}{-1}%
  \endgroup
}

\newcounter{finding}

\newcommand{\finding}[1]{%
  \refstepcounter{finding}%
  \begin{tcolorbox}[
        colback=white!90!gray,
        colframe=teal!60!black,
        arc=5pt,
        boxsep=5pt,
        left=1mm,
        right=1mm,
        top=1mm,
        bottom=1mm,
        boxrule=0.8pt,
        drop shadow=gray!50!white,
        enhanced jigsaw
    ]
    \vspace{-0.1cm}
        \paragraph{\textbf{\textit{Finding \thefinding:}}} #1
        \label{finding:\thefinding}
    \vspace{-0.1cm}
    \end{tcolorbox}
    \vspace{-0.1cm}
}

\title{\texttt{code\_transformed}: The Influence of Large Language Models on Code}

\author{
 \textbf{Yuliang Xu\textsuperscript{1\textdagger}}, 
 \textbf{Siming Huang\textsuperscript{1\textdagger}},
 \textbf{Mingmeng Geng\textsuperscript{2 3 4*}},
 \textbf{Yao Wan\textsuperscript{1*}}, 
 \textbf{Xuanhua Shi\textsuperscript{1}}, 
 \textbf{Dongping Chen\textsuperscript{1\textdaggerdbl}}\\
 \textsuperscript{1} Huazhong University of Science and Technology, 
 \textsuperscript{2} École normale supérieure - Université PSL, \\ 
 \textsuperscript{3} Laboratoire LATTICE,
 \textsuperscript{4} International School for Advanced Studies (SISSA)\\
 \texttt{mingmeng.geng@ens.psl.eu}, \texttt{wanyao@hust.edu.cn}
 }

\begin{document}
\maketitle
\blfootnote{\textdagger \ Equal Contribution. * Corresponding Authors. \textdaggerdbl \ Project  Lead.}

\begin{abstract}
Coding remains one of the most fundamental modes of interaction between humans and machines. With the rapid advancement of \textit{Large Language Models} (LLMs), code generation capabilities have begun to significantly reshape programming practices. This development prompts a central question: \textit{Have LLMs transformed code style, and how can such transformation be characterized?} In this paper, we present a pioneering study that investigates the impact of LLMs on code style, with a focus on naming conventions, complexity, maintainability, and similarity. By analyzing code from over 20,000 GitHub repositories linked to arXiv papers published between 2020 and 2025, we identify measurable trends in the evolution of coding style that align with characteristics of LLM-generated code. For instance, the proportion of \textit{snake\_case} function names in Python code increased from 40.7\% in Q1 2023 to 49.8\% in Q3 2025. Furthermore, we investigate how LLMs approach algorithmic problems by examining their reasoning processes. Our experimental results may provide the first large-scale empirical evidence that LLMs affect real-world programming style.\footnote{We release all the experimental dataset and source code at: \href{https://github.com/ignorancex/LLM_code}{https://github.com/ignorancex/LLM\_code}.}

\end{abstract}

\section{Introduction}
The widespread use of Large Language Models (LLMs) and LLM-powered tools (such as Copilot and Cursor) in coding has been confirmed through questionnaire surveys~\citep{liang2024large,daigle2024survey,peslak2024ai,li2025engineering}. Our paper takes a different perspective: \textit{Can the transformation driven by LLMs be observed in real-world code?} 

\begin{figure}[!t]
    \centering
    \begin{subfigure}[b]{\linewidth}
        \centering
        \includegraphics[width=.98\linewidth]{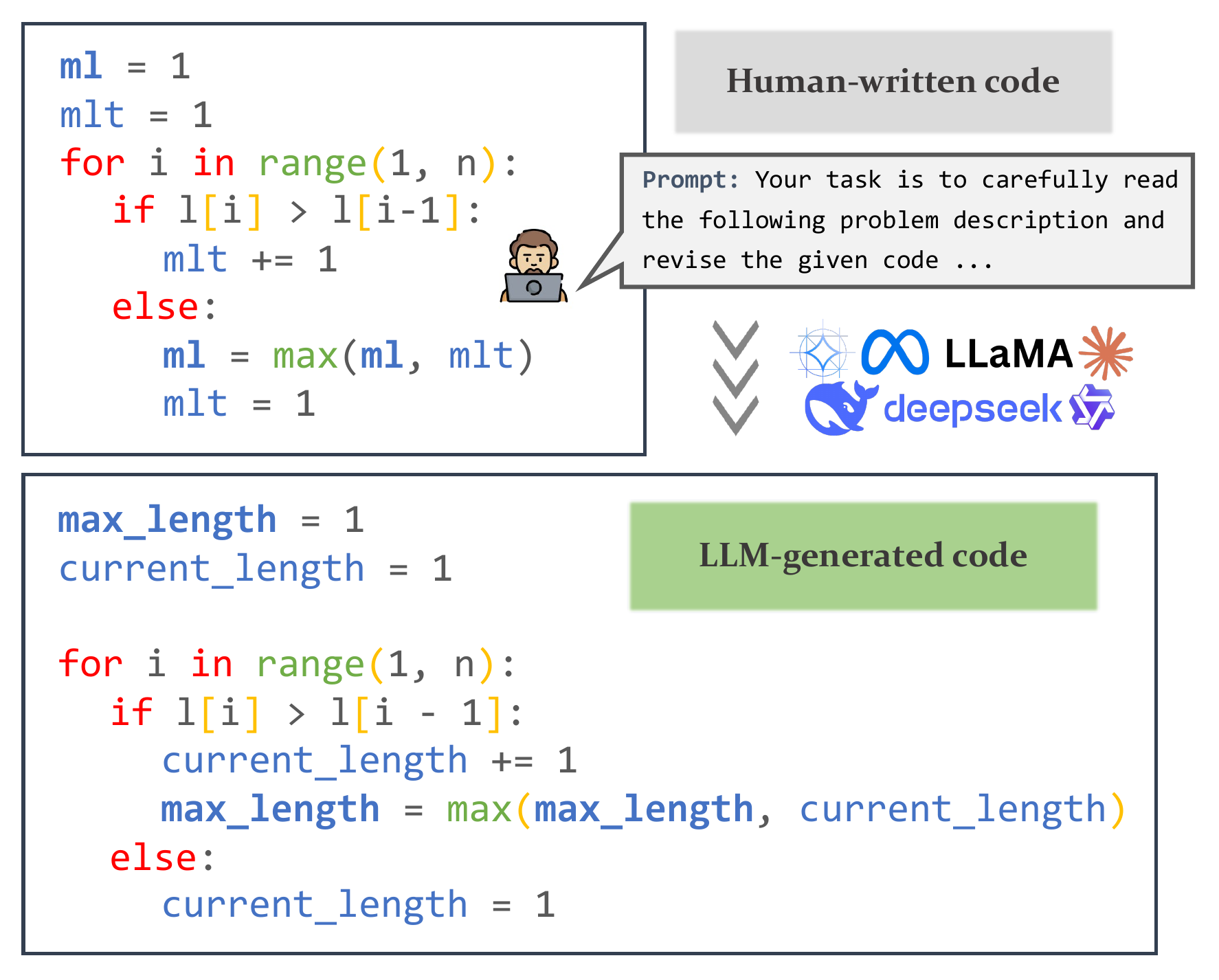}
        \caption{An example of code rewritten by LLMs.}
        \label{llm_example}
    \end{subfigure}

    \vspace{0.5em}

    \begin{subfigure}[b]{\linewidth}
        \centering

        \begin{subfigure}[b]{0.49\linewidth}
            \centering
            \includegraphics[width=\linewidth]{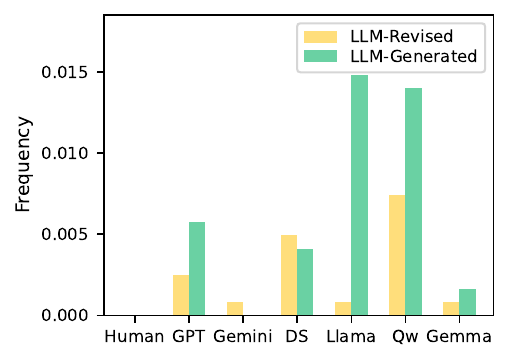}
        \end{subfigure}
        \hfill
        \begin{subfigure}[b]{0.49\linewidth}
            \centering
            \includegraphics[width=\linewidth]{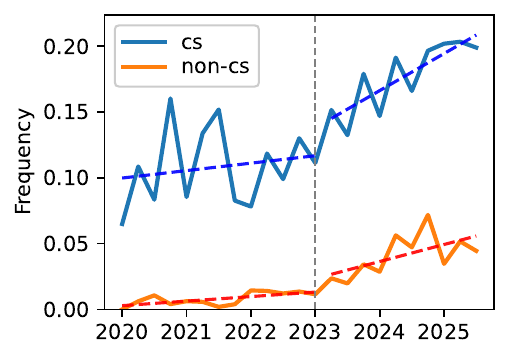}
        \end{subfigure}
        \caption{Frequency of the variable name \texttt{max\_length} in our human-written code dataset and LLM-processed code, as well as its coverage across the GitHub repositories we collected. The dashed lines represent the results of the linear regressions.}
        \label{var_example}
    \end{subfigure}
    \caption{LLMs’ preference for longer variable names and \textit{snake\_case} naming patterns.}
    \label{overview}
\end{figure}

Previous research has explored the influence of LLMs in the text and speech domains~\citep{liang2024monitoring,geng2024impact}. Code generated by LLMs differs in style from human-written code~\citep{wang2024beyond,park2025detection}.  Thus, we could potentially adopt a similar approach to observe the impact of LLMs on code. Figure~\ref{llm_example} presents a motivating example where the short variable names (i.e., \texttt{ml} and \texttt{mlt}) are replaced with longer and more descriptive names following the \textit{snake\_case} convention (i.e., \texttt{max\_length} and \texttt{current\_length}). In addition, Figure~\ref{var_example} shows that LLMs are more likely to use the variable name \texttt{max\_length} compared to humans, and its prevalence in GitHub repositories has also increased.

The rapid adoption of LLMs has raised concerns about code integrity~\cite{wadhwa2024core}, potential copyright infringement~\cite{wan2024does}, and broader ethical or legal implications~\cite{xu2024licoeval}. These issues have motivated efforts to trace and attribute the influence of LLM-assisted programming. Therefore, we conduct a pioneering study to investigate the influence of LLMs on code from the following perspectives.

From the perspective of \textbf{naming patterns}, we first categorize the names of variables, functions, and files into several distinct formats (e.g. \textit{snake\_case}). By analyzing the names of variables and functions in the GitHub repository code, we observe a clear increase in the usage of LLM-preferred naming styles.

From the perspective of \textbf{code complexity} and \textbf{maintainability}, we simulate code generated and rewritten by LLMs, extract subsets for analysis, and compare them with human-written code from GitHub. Our results indicate that LLM-rewritten code tends to be more concise under certain metrics—notably, cyclomatic complexity in Python, although this improvement is less pronounced in stylistic aspects such as naming conventions and no clear trend is observed in the GitHub code. 

From the perspective of \textbf{code similarity}, the rewritten code exhibits relatively high similarity to the original, especially compared to code generated directly by LLM. This observation further highlights that different usage scenarios, such as code rewriting versus direct generation, can produce different outcomes. 

Based on the above findings, we further explore whether the generated content influences the subsequent code generation abilities of LLMs. By analyzing the models’ thinking process, we find that their outputs do not always align with the expected algorithmic approaches for the given problems.

We believe the findings in our work will enhance knowledge about LLMs' programming abilities and coding styles, providing novel insights for assessing and monitoring their broader impacts. 
\section{Background}
\label{metrics}

\subsection{Position of Our Work}

Comparisons between code generated by LLMs and that written by humans can be conducted from multiple perspectives. Previous research has investigated various methods to distinguish between the two, such as using perplexity scores~\citep{xu2024detecting} and manually designed features~\citep{bulla2024ex,park2025detection}. 

However, rather than focusing on differentiating LLM-generated code from human-written code, our work considers a more realistic and increasingly common scenario: LLM-assisted code authoring, where human developers and language models collaboratively produce code. For example, we compare the similarity between human-written code and code that has been rewritten or generated by LLMs using cosine similarity and Jaccard similarity. In order to better grasp how LLMs generate code, we also carefully analyze the reasoning chains to see if they think about the algorithms that the problems were designed to elicit.

\begin{figure*}[!t]
    \centering
    \includegraphics[width=\linewidth]{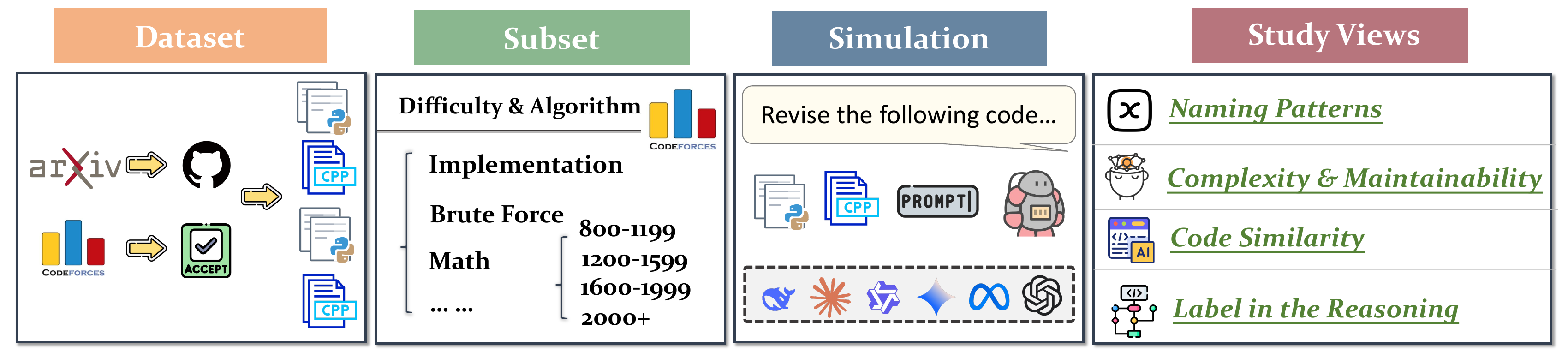}
    \caption{The process of our experiments.}
    \vspace{-1em}
    \label{framework}
\end{figure*}

\subsection{Code Style Measurements}
\label{tasks}
\paragraph{Naming Patterns.} The naming in code generated by LLMs has its own characteristics~\citep{park2025detection}. Therefore, we categorize variable, function, and file names into several distinct formats: \textit{single letter}, \textit{lower-case}, \textit{UPPERCASE}, \textit{camelCase}, \textit{snake\_case}, \textit{PascalCase}, and \textit{endsWithDigits}. Besides, the length of the names is also considered. 

\paragraph{Cyclomatic Complexity.} 
Cyclomatic complexity~\citep{mccabe1976complexity} is a metric used to measure the number of linearly independent paths in a program's control flow. \citet{graylin2009cyclomatic} explored the relationship between cyclomatic complexity and lines of code. Some researchers use this method to analyze code generated by LLMs~\citep{dou2024s}. 
Given the control-flow graph (CFG) of a code snippet, let $E$ denote the number of edges, $N$ the number of nodes, and $P$ the number of connected components. The cyclomatic complexity is calculated by \( G = E - N + 2P \).  
For a single connected component (\( P = 1 \)), it simplifies to the number of decision points plus one. Each occurrence of \texttt{if}, \texttt{for}, \texttt{while}, etc., is counted as one decision point.

\paragraph{Halstead Complexity Metrics.} Halstead complexity provides a quantitative assessment of code complexity based on the use of operands and operators \citep{hariprasad2017software}. It can be viewed from the following perspectives:

\begin{itemize}[leftmargin=*,itemsep=0pt]
  \item Program Vocabulary: \( n = n_1 + n_2 \)
  \item Program Length: \( N = N_1 + N_2 \)
  \item Calculated Program Length: \( \hat{N} = n_1 \cdot \log_2 n_1 + n_2 \cdot \log_2 n_2 \)
  \item Volume: \( V = N \cdot \log_2 n \)
  \item Difficulty: \( D = \left( \frac{n_1}{2} \right) \cdot \left( \frac{N_2}{n_2} \right) \)
  \item Effort: \( E = D \cdot V \)
  \item Time to Implement: \( T = \frac{E}{18} \) (in seconds)
  \item Estimated Bugs: \( B = \frac{E^{2/3}}{3000} \)
\end{itemize}
where \( n_1 \) is the number of distinct operators,  
\( n_2 \) is the number of distinct operands,  
\( N_1 \) is the total number of operator occurrences, and  
\( N_2 \) is the total number of operand occurrences.

\paragraph{Maintainability Index.}  The metrics related to maintainability are defined as follows:

\begin{itemize}[leftmargin=*,itemsep=0pt]
  \item Standard maintainability index:
  \[
  \text{MI}_{\text{std}} = 171 - 5.2 \cdot \ln(V) - 0.23 \cdot \text{CC} - 16.2 \cdot \ln(\text{S})
  \]
  
  \item Custom maintainability index:
  \[
  \text{MI}_{\text{custom}} = \text{MI}_{\text{std}} + 50 \cdot \sin\left(\sqrt{2.4 \cdot \text{CR}}\right)
  \]
\end{itemize}
where \( V \) denotes the Halstead volume, \( \text{CC} \) is the cyclomatic complexity, \( \text{S} \) is the number of logical source lines of code, and \( \text{CR} \) is the comment ratio.

\paragraph{Code Similarity.}

Let $A$ and $B$ be the words of the code segment, and let $\vec{v}_A$ and $\vec{v}_B$ be their corresponding vector representations. Then we can use the following cosine similarity to compare the similarity of the code, defined as follows:
\begin{equation}
    \text{sim}_{\text{cosine}}(A, B)= \frac{\vec{v}_A \cdot \vec{v}_B}{\|\vec{v}_A\| \cdot \|\vec{v}_B\|} \,.
\end{equation}
Similarly for Jaccard similarity:
\begin{equation}
\text{sim}_{\text{J}}(A, B) =
\begin{cases}
&1,  \text{ if } A = \emptyset \text{ and } B = \emptyset \\
&0,  \text{ if } A = \emptyset \text{ or } B = \emptyset \\ &
\frac{|A \cap B|}{|A \cup B|},  \text{ otherwise}
\end{cases} \,
\end{equation}

\paragraph{Label Similarity.} 
To refine our analysis, we analyze the matching of reasoning and labels for each question separately. Let \( T \) denote the set of all labels. For each question \( q \), let \( A_q \subseteq T \) denote the set of ground-truth labels (from the question description), and let \( R_q \subseteq T \) be the set of labels appearing in the reasoning process. Then we define the $\mathrm{match}$ and $\mathrm{error}$ metrics as follows:
\begin{align}
    \operatorname{match}(q) &= \mathbf{1}\{ A_q \cap R_q \neq \emptyset \}, \\
    \operatorname{error}(q) &= \mathbf{1}\{ (T \setminus A_q) \cap R_q \neq \emptyset \},
\end{align}
where \(\mathbf{1}\{\cdot\}\) is the indicator function, i.e., \(\mathbf{1}\{P\} = 1\) if condition \(P\) is true and \(0\) if condition \(P\) is false. The match rate and error rate are computed as the average number of matches and errors per question.

\section{Study Design}
As illustrated in Figure~\ref{framework}, our process begins with the collection of human-written code from GitHub and Codeforces, after which we generate code using LLMs with various prompting strategies. By comparing the differences between human-written and LLM-generated solutions, and analyzing the temporal trends of these metrics on GitHub, we investigate the relationship between the two. Furthermore, to broaden the scope of our study, we select a subset of problems and evaluate code generated by a wider range of models to explore stylistic variations across LLM-generated code.

\subsection{Dataset}

\paragraph{Human-Written Code.}
We utilize \texttt{Code4Bench}, a multidimensional benchmark based on Codeforces data~\citep{majd2019code4bench}. This dataset contains user submissions on Codeforces before 2020, which were barely impacted by LLMs. 

\paragraph{GitHub Data.} We collect a total of 22,074 GitHub repositories and 1,032,016 source code files, corresponding to arXiv papers from the first quarter of 2020 to the third quarter of 2025. In order to reduce the impact of confounding factors, each repository in the dataset is labeled with two attributes: the programming language, which can be either Python or C/C++, and the scientific domain, indicating whether the associated arXiv paper belongs to, for example, computer science (cs) and non-computer science (non-cs). The number of repositories and files of each language per quarter based is shown in Table~\ref{tab:quarter_repo_stats}.

\paragraph{Problem Subset.}

To reduce computational costs while maintaining representativeness, we select 200 questions from \texttt{Code4Bench}, spanning a range of difficulty levels and algorithm types and categorize them into four groups based on their difficulty rating: 800–1199, 1200–1599, 1600–1999, and 2000+. The first valid tag is utilized to determine each problem's primary algorithm. We then filter problems that are annotated with one of the following ten target algorithms: \textit{implementation, brute force, constructive algorithms, greedy, binary search, math, dp, data structures, combinatorics,} and \textit{dfs and similar}. From each difficulty group, we randomly sample 50 problems. The number of problems per algorithm in our sample approximates its distribution in the original group. Detailed information is provided in table \ref{subset}.

\subsection{Studied LLMs}
\label{models}

A diverse set of LLMs is used to cover different architectures and scales. The \textit{Qwen3} series (4B, 8B, 14B, 32B) and \textit{Qwen2.5-Coder-32B-Instruct} are chosen for exploring the impact of model sizes on code generation \citep{qwen3,hui2024qwen2}. The DeepSeek family, including \textit{DeepSeek-V3}, \textit{DeepSeek-R1}, and \textit{DeepSeek-R1-Distill-Qwen-32B}, was used to evaluate the reasoning performance~\citep{deepseekai2024deepseekv3technicalreport,deepseekai2025deepseekr1incentivizingreasoningcapability}. We also incorporate leading general-purpose LLMs, \textit{GPT-4o-mini} \citep{hurst2024gpt}, \textit{GPT-4.1} \citep{openai2025gpt41}, \textit{Claude-3.7-Sonnet} \citep{claude-3-5}, \textit{Gemma-3-27B} \citep{team2025gemma}, \textit{Gemini-2.0-flash} \citep{google2024gemini20}, \textit{Llama-3.3-Nemotron-Super-49B-V1} \citep{bercovich2025llamanemotronefficientreasoningmodels}, and \textit{Llama-4-Maverick} \citep{meta2025llama} to ensure broad coverage of both closed- and open-source training paradigms.

\subsection{Simulations}
\label{generation_strategies}
We apply two code generation strategies across six LLMs: \textit{GPT-4.1}, \textit{Gemini-2.0-flash}, \textit{Deepseek-R1-Distill-Qwen-32B}, \textit{Llama-4-Maverick}, \textit{Qwen3-32B}, and \textit{Gemma-3-27B}. The prompts for the two strategies are shown in Figures \ref{prompt_generate_1} and \ref{prompt_revise}.

\paragraph{Direct Generation.} LLMs are provided only with the problem description and asked to generate a solution from scratch. 

\paragraph{Reference-Guided Generation.} In addition to the problem description, the model is also given a reference solution (i.e., a user-submitted, passed code). The model is instructed to analyze this code and revise it when generating its solution.

\paragraph{Syntax Validation.} For C/C++, we compile both direct-generated and reference-guided code using \texttt{g++ -fsyntax-only}. For Python, we performed static analysis with \texttt{ast.parse}. Only samples where both code blocks passed syntax checks were retained, ensuring syntactic correctness.

\subsection{Evaluation}
In addition to examining the similarities and differences between human-written code and LLM-generated code, we also aim to compare different LLMs and assess whether some models produce code that more closely resembles human style. To this end, we conduct a larger-scale evaluation across a broader set of models.
Based on our subset, we expand our analysis to all models listed in Section~\ref{models}, excluding \textit{DeepSeek-R1-Distill-Qwen-32B}, \textit{GPT-4.1}, \textit{Llama-4-Maverick}, and \textit{Gemini-2.0-flash}. For each problem, we prompt the models to generate code using the template in Figure~\ref{prompt_generate_2}, repeating this process 32 times.

\begin{figure*}[!t]
    \centering
    \begin{subfigure}[t]{0.24\linewidth}
        \centering
        \includegraphics[width=\linewidth]{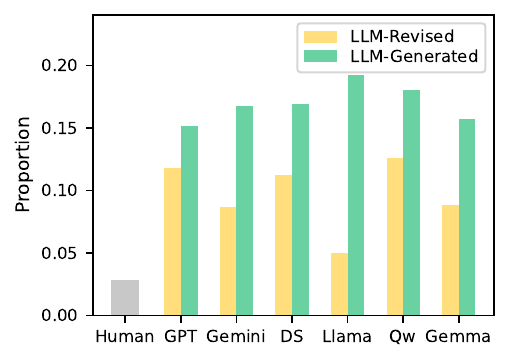}
        \caption{\textit{snake\_case} variables.}
        \label{cf_py_var_snake}
    \end{subfigure}
    \hfill
    \begin{subfigure}[t]{0.24\linewidth}
        \centering
        \includegraphics[width=\linewidth]{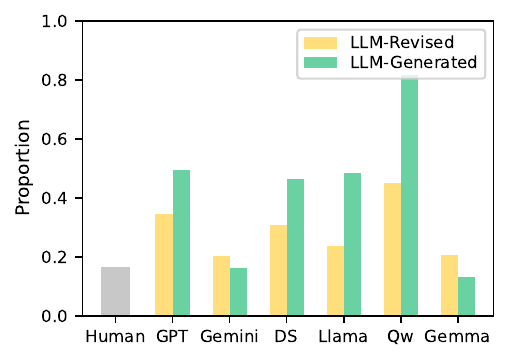}
        \caption{\textit{snake\_case} functions.}
        \label{cf_py_func_snake}
    \end{subfigure}
    \hfill
    \begin{subfigure}[t]{0.24\linewidth}
        \centering
        \includegraphics[width=\linewidth]{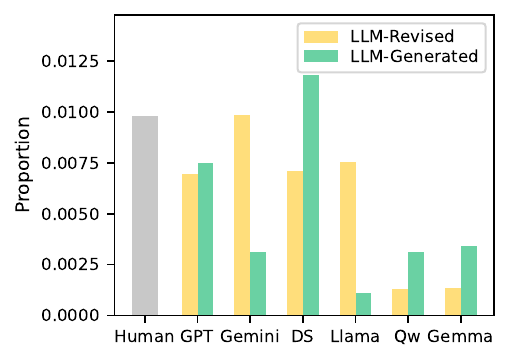}
        \caption{Digit-suffixed functions.}
        \label{cf_py_func_digit}
    \end{subfigure}
    \hfill
    \begin{subfigure}[t]{0.24\linewidth}
        \centering
        \includegraphics[width=\linewidth]{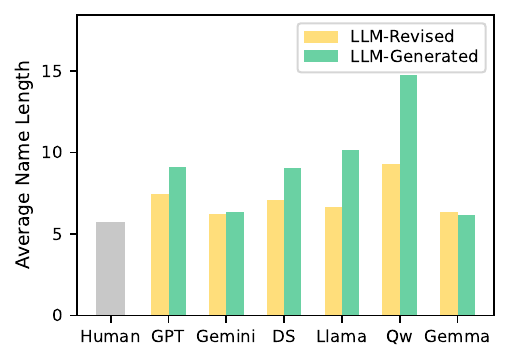}
        \caption{Length of functions.}
        \label{cf_py_func_len}
    \end{subfigure}

    \begin{subfigure}[t]{0.24\linewidth}
        \centering
        \includegraphics[width=\linewidth]{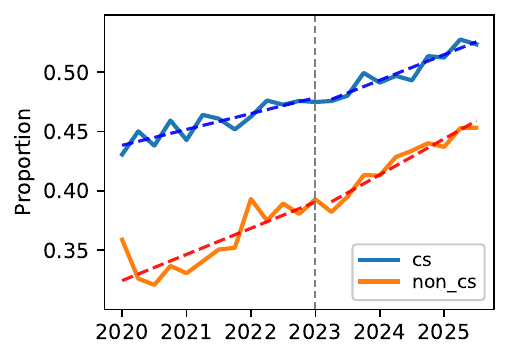}
        \caption{\textit{snake\_case} variables.}
        \label{py_var_snake}
    \end{subfigure}
    \hfill
    \begin{subfigure}[t]{0.24\linewidth}
        \centering
        \includegraphics[width=\linewidth]{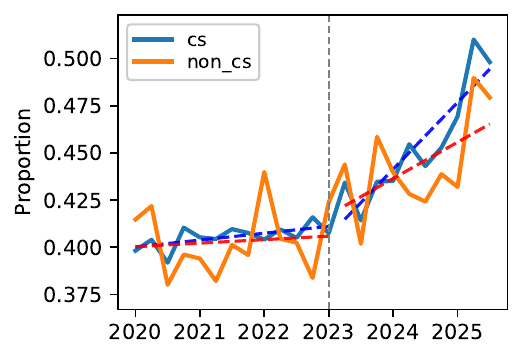}
        \caption{\textit{snake\_case} functions.}
        \label{py_func_snake}
    \end{subfigure}
    \hfill
    \begin{subfigure}[t]{0.24\linewidth}
        \centering
        \includegraphics[width=\linewidth]{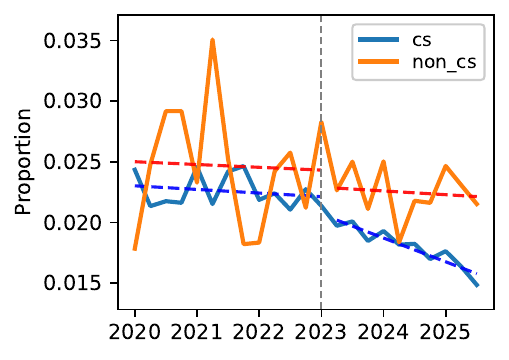}
        \caption{Digit-suffixed functions.}
        \label{py_func_digit}
    \end{subfigure}
    \hfill
    \begin{subfigure}[t]{0.24\linewidth}
        \centering
        \includegraphics[width=\linewidth]{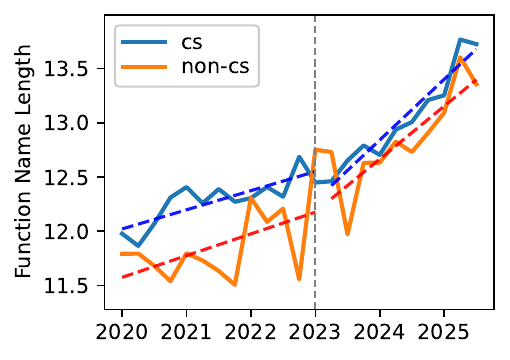}
        \caption{Length of functions.}
        \label{py_func_len}
    \end{subfigure}   

    \caption{The four figures in the first row present simulation results derived from Codeforces human-written code, either revised by LLMs or directly generated by LLMs based on problem descriptions. The four figures in the second row show the trends over time in GitHub repositories for Python variable names using \textit{snake\_case}, digit-suffixed function names, and the length of variable names. Model abbreviations: GPT (\textit{GPT-4.1}), Gemini (\textit{Gemini-2.0-flash}), DS (\textit{DeepSeek-R1-Distill-Qwen-32B}), Llama (\textit{Llama-4-Maverick}), Qw (\textit{Qwen3-32B}), Gemma (\textit{Gemma-3-27B}).
    }
    \vspace{-1em}
\end{figure*}

\section{View I: Naming Patterns}
\subsection{Settings}

We consider the following formats of names: \textit{single letter}, \textit{lowercase}, \textit{UPPERCASE}, \textit{camelCase}, \textit{snake\_case}, \textit{PascalCase}, and \textit{endsWithDigits}. Any name that does not match these specific patterns is grouped into the \textit{Other} category. The length of the names is regarded as an additional metric.

To extract names from source code, we apply different strategies depending on the programming language. For Python code, we use the \texttt{ast} module to statically parse the abstract syntax tree and extract function and variable names. For C/C++ code, we use regular expressions to identify name patterns directly from the source text. All extracted names are then matched against predefined regular expressions to classify them into the aforementioned naming formats.

To prevent large repositories from dominating the overall distribution, we normalize at the repository level: we first compute naming pattern distributions for each file, then average these distributions within each repository, and finally average across repositories to obtain the overall statistics.

\subsection{Results}

\paragraph{Patterns in LLM-generated Code.} LLMs have slight deviations from general human naming conventions when it comes to variables and functions. For instance, Figures~\ref{cf_py_var_snake} and \ref{cf_py_func_snake} illustrate that all three evaluated LLMs tend to use \textit{snake\_case} in names compared to human-written code. Figure~\ref{cf_py_func_len} shows that LLMs tend to use longer function names. But there isn't always a clear-cut, such as the \textit{digit-suffixed} naming pattern plotted in Figure~\ref{cf_py_func_digit}.  

\paragraph{Trends in GitHub Repositories.} Figures \ref{py_var_snake} and \ref{py_func_snake} show the adoption of \textit{snake\_case} names steadily rises in both CS and non-CS projects, which is consistent with the stylistic differences observed between human-written code and LLM-generated code. Similarly, Figure~\ref{py_func_len} presents the growth in the length of function names in GitHub code. Although some trends had already emerged before LLM era, it's likely that LLMs have accelerated the process. We are also considering further subdividing the categories, and the trends may vary, as shown in Figures~\ref{detail_var} and~\ref{detail_func} in the appendix.

\paragraph{Influence of Disciplines.} For non-CS repositories, naming patterns sometimes exhibit greater fluctuation compared to the clearer trends observed in CS repositories. A good example is the trend of Python function names ending with digits, as shown in Figure~\ref{py_func_digit}. While CS repositories demonstrate a steady decline in the use of such names, non-CS projects show more variability. 

\paragraph{Differences between Programming Languages.} Unlike Python, fewer naming patterns show clear temporal trends in C/C++ repositories. However, there are still notable cases: the use of \textit{snake\_case} in both variable and function names shows an upward trend as shown in Figures \ref{cpp_var_snake} and \ref{cpp_func_snake}, while the use of lowercase names in variables declines over time. Both of them align with the stylistic tendencies observed in LLM-generated code. 

\paragraph{Other Evidences.}  More results are provided in Appendix~\ref{github naming}. For example, Figures~\ref{python_var_digit} and~\ref{python_var_single} show that most LLMs tend to avoid \textit{digit-suffixed} and \textit{single-letter} variable names, and these patterns also steadily decline in both CS and non-CS repositories. Similar trends are observed in the function names.  These parallel developments suggest a potential correlation between human and machine-generated coding styles.

\begin{figure*}[!t]
    \centering
    \begin{subfigure}[t]{0.48\linewidth}
        \centering
        \includegraphics[width=\linewidth]{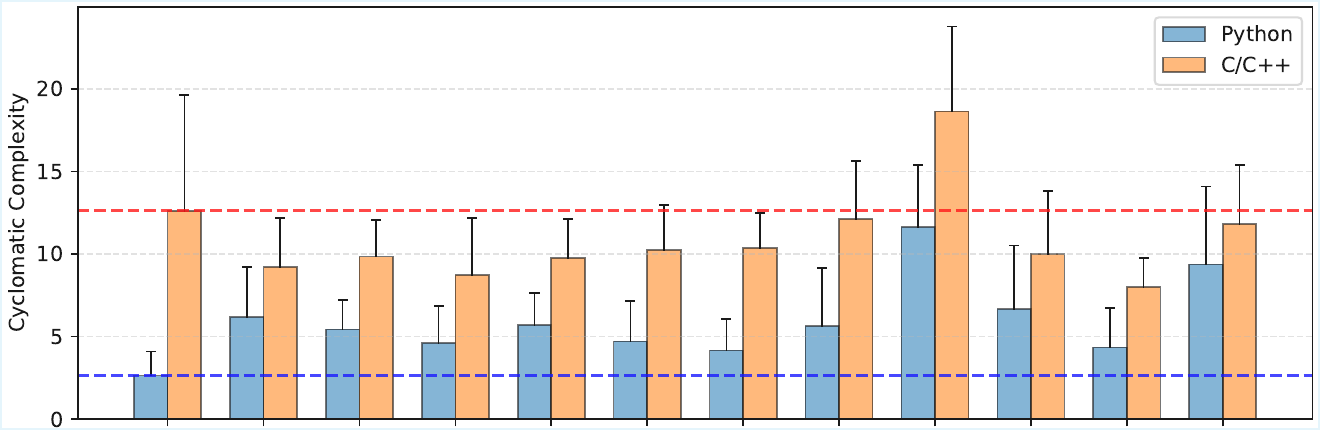}
        \caption{Cyclomatic Complexity.}
        \label{error_bar_cyc}
    \end{subfigure}
    \hfill
    \begin{subfigure}[t]{0.48\linewidth}
        \centering
        \includegraphics[width=\linewidth]{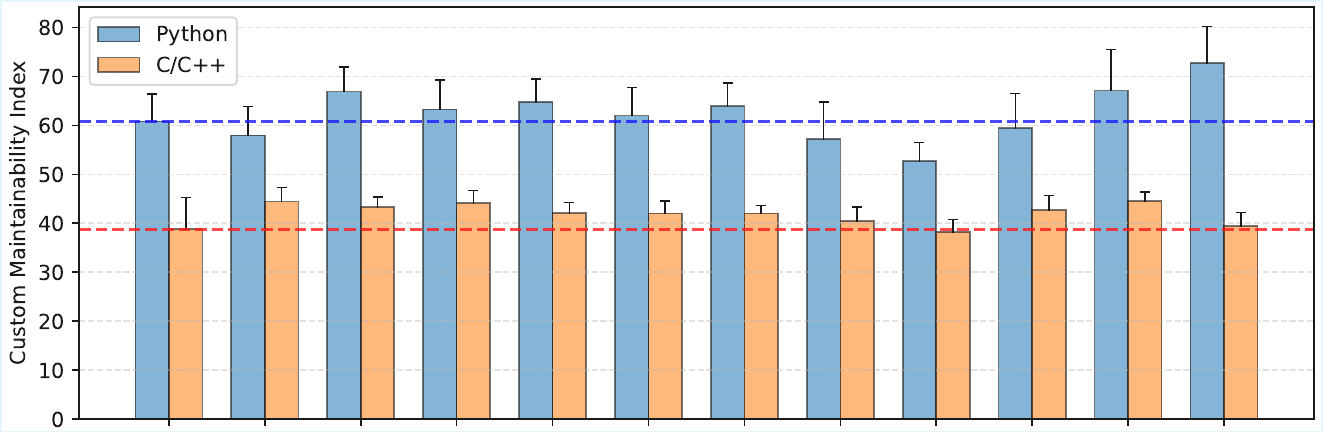}
        \caption{Custom Maintainability Index.}
        \label{error_bar_maintain}
    \end{subfigure}

    \begin{subfigure}[t]{0.48\linewidth}
        \centering
        \includegraphics[width=\linewidth]{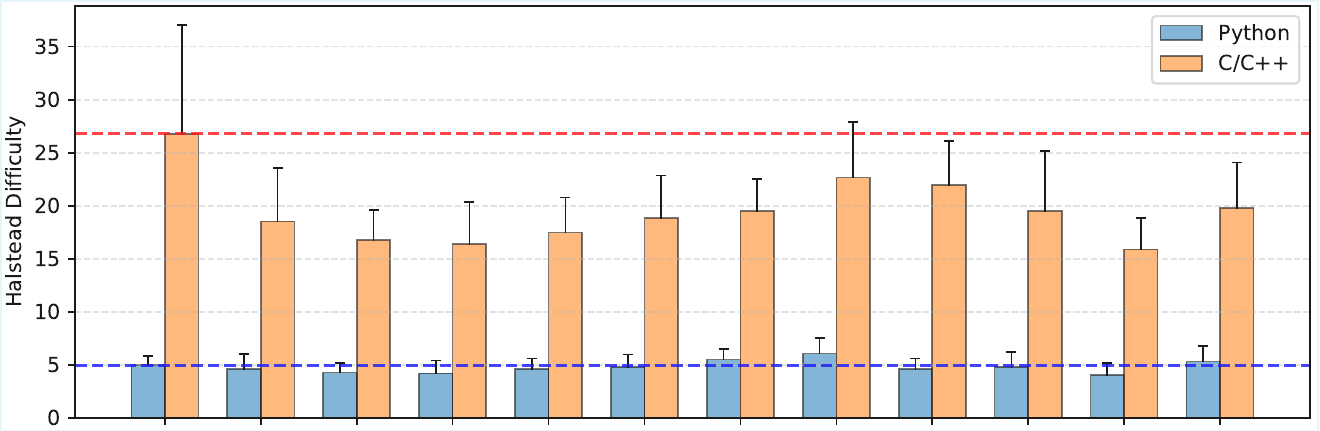}
        \caption{Halstead Difficulty.}
        \label{error_bar_diff}
    \end{subfigure}
    \hfill
    \begin{subfigure}[t]{0.48\linewidth}
        \centering
        \includegraphics[width=\linewidth]{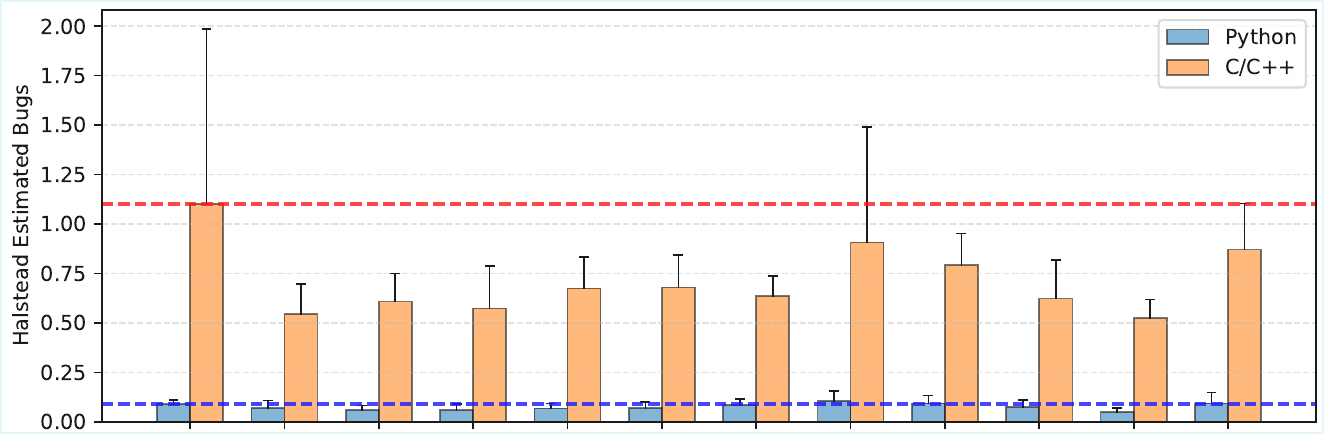}
        \caption{Halstead Estimated Bugs.}
        \label{error_bar_bugs}
    \end{subfigure}

    \begin{subfigure}[t]{0.24\linewidth}
        \centering
        \includegraphics[width=\linewidth]{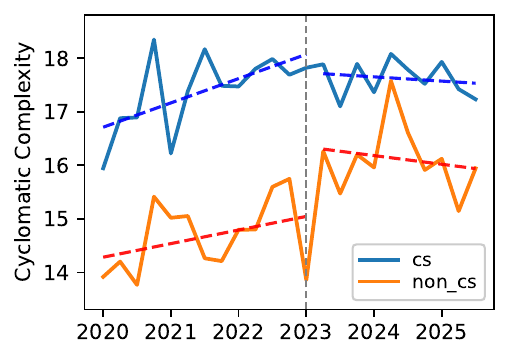}
        \caption{Cyclomatic Complexity.}
        \label{git_cyclo}
    \end{subfigure}
        \hfill
    \begin{subfigure}[t]{0.24\linewidth}
        \centering
        \includegraphics[width=\linewidth]{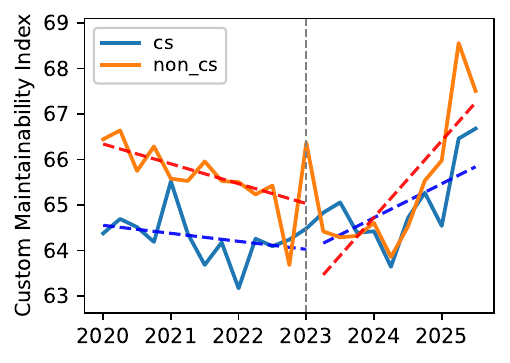}
        \caption{Maintainability Index.}
        \label{git_maintain}
    \end{subfigure}
    \begin{subfigure}[t]{0.24\linewidth}
        \centering
        \includegraphics[width=\linewidth]{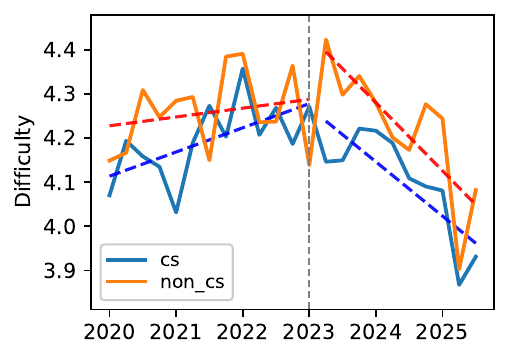}
        \caption{Halstead Difficulty.}
        \label{git_diff}
    \end{subfigure}
        \hfill
    \begin{subfigure}[t]{0.24\linewidth}
        \centering
        \includegraphics[width=\linewidth]{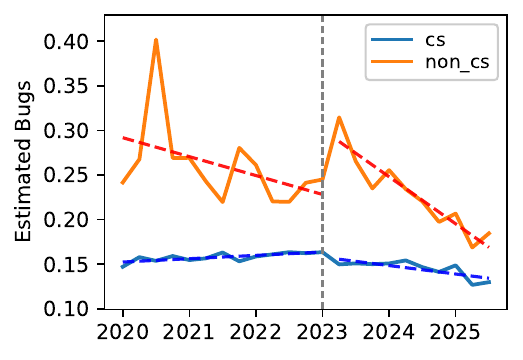}
        \caption{Halstead estimated Bugs.}
        \label{git_bugs}
    \end{subfigure}

    \caption{(a)--(d) present the results of code generated by various models on our subset problems evaluated using four metrics. The leftmost bars correspond to human-written code. The evaluated models, from left to right, are \textit{Qwen2.5-Coder-32B-Instruct}, \textit{Qwen3-4B}, \textit{Qwen3-8B}, \textit{Qwen3-14B}, \textit{Qwen3-32B}, \textit{DeepSeek-V3}, \textit{DeepSeek-R1}, \textit{Gemma-3-27B}, \textit{Llama-3.3-Nemotron-Super-49B-V1}, \textit{GPT-4o-mini}, and \textit{Claude-3.5-Sonnet}. (e)--(h) illustrate the evolution of Python code on GitHub according to the same metrics.}

    \label{metrics_subset_figure}
\end{figure*}

\finding{The coding style of human-written code may be influenced by LLMs: they may not only mirror existing norms but also subtly reshape them, gradually pushing human developers toward greater stylistic alignment with LLM-preferred conventions.}

\section{View II: Complexity and Maintainability} 
\subsection{Settings}
In order to explore the difference between code written by humans and LLM, we assess them from multiple dimensions, including information volume, control flow complexity, code structure, and adequacy of comments.
We first calculate the mean and standard deviation for each problem across 32 generated outputs per model, using metrics defined in Section~\ref{metrics}. Subsequently, we average these per-problem statistics across all problems to obtain final results for each model. These final values reflect each model's overall performance on the generation task, as well as the stability of its output for individual problems.

\begin{table*}[!t]
\small
\renewcommand{\arraystretch}{1.2}
\setlength{\tabcolsep}{6pt}
\centering
\begin{tabular}{l|ccc|ccc}
\toprule
\multirow{2}{*}{\textbf{Model}} 
& \multicolumn{3}{c|}{\textbf{C/C++}} 
& \multicolumn{3}{c}{\textbf{Python}} \\
\cmidrule(lr){2-4} \cmidrule(lr){5-7}
& AC vs ANS & AC vs REF & ANS vs REF
& AC vs ANS & AC vs REF & ANS vs REF \\
\midrule
Qwen3-32B     & 0.2805 & 0.6370 & 0.3686 & 0.2103 & 0.5629 & 0.2936 \\
Gemma-3-27B    & 0.2993 & 0.7417 & 0.3941 & 0.2514 & 0.7202 & 0.3279 \\
Deepseek-R1-Distill-Qwen-32B & \textbf{0.3172} & 0.7375 & 0.3990 & 0.2587 & 0.5047 & 0.3685 \\
GPT-4.1      & 0.2874 & 0.5281 & \textbf{0.4220} & 0.2468 & 0.5033 & \textbf{0.3829} \\
Gemini-2.0-Flash   & 0.3149 & 0.7440 & 0.3943 & \textbf{0.2744} & 0.7697 & 0.3350 \\
Llama-4-Maverick   & 0.2989 & \textbf{0.8005} & 0.3508 & 0.2274 & \textbf{0.8486} & 0.2644 \\
\bottomrule
\end{tabular}
\caption{Cosine similarity between the human-written code (AC), the initial LLM-generated code (ANS), and the LLM-rewritten code (REF) based on human-written code.}
\label{cos}
\end{table*}

\begin{table*}[!t]
\small
\renewcommand{\arraystretch}{1.2}
\setlength{\tabcolsep}{6pt}
\centering
\begin{tabular}{l|ccc|ccc}
\toprule
\textbf{Model} 
& \multicolumn{3}{c|}{\textbf{C/C++}} 
& \multicolumn{3}{c}{\textbf{Python}} \\
\cmidrule(lr){2-4} \cmidrule(lr){5-7}
& AC vs ANS & AC vs REF & ANS vs REF
& AC vs ANS & AC vs REF & ANS vs REF \\
\midrule
Qwen-32B     & 0.2955 & 0.6173 & 0.3852 & 0.2885 & 0.6079 & 0.3518 \\
Gemma-3-27B    & 0.3134 & 0.7391 & 0.4171 & 0.3742 & 0.7549 & 0.4451 \\
Deepseek-R1-Distill-Qwen-32B & 0.3250 & 0.7034 & 0.4161 & 0.3705 & 0.5554 & \textbf{0.4626} \\
GPT-4.1      & 0.2649 & 0.4546 & 0.3787 & 0.3308 & 0.5425 & 0.4376 \\
Gemini-2.0-Flash   & \textbf{0.3258} & 0.7370 & \textbf{0.4214} & \textbf{0.4056} & 0.8062 & 0.4618 \\
Llama-4-Maverick   & 0.3160 & \textbf{0.7513} & 0.3874 & 0.3514 & \textbf{0.8404} & 0.3895 \\
\bottomrule
\end{tabular}
\caption{Jaccard similarity between the human-written code (AC), the initial LLM-generated code (ANS), and the LLM-rewritten code (REF) based on human-written code.}
\label{jaccard}
\end{table*}

\subsection{Results}

\paragraph{LLM-Generated \textit{vs} LLM-Revised.}  Tables~\ref{tab:halstead_stats} and~\ref{tab:maintainability_stats} present the results containing direct generation and reference-guided generation. For the same model in Python, the Halstead volume and effort of the solution obtained using the accepted (AC) code as the reference (REF) are generally lower than those generated directly (ANS). For example, Deepseek-REF-Python exhibits lower volume (195.18 \textit{vs.} 201.03) and effort (1345.86 vs. 1472.81) than Deepseek-ANS-Python. But the opposite trend is observed in C/C++, suggesting that for Python, reference-guided generation leads to more concise code, whereas in C/C++, direct generation yields more concise results. 

\paragraph{LLM \textit{vs} Human.} 
Figure~\ref{error_bar_cyc} illustrates the cyclomatic complexity of the code written by human and different models. Compared to human-written code, all LLMs except for \textit{Gemma3-27B} generated C/C++ code with lower cyclomatic complexity. However, for Python, the code produced by all models is more complex than human-written one.
This may suggest that LLMs are more accustomed to solving algorithmic problems in C/C++ and can produce high-quality solutions in that language, but tend to underperform when using Python. Figure~\ref{error_bar_maintain} shows that human-written code exhibits the second-lowest maintainability in Python and the lowest in C/C++, suggesting that LLM-generated code tends to be easier to maintain. Moreover, Figures~\ref{error_bar_diff} and~\ref{error_bar_bugs} demonstrate that LLMs can produce code with lower Halstead difficulty and fewer bugs.
More results can be found in Tables~\ref{tab:halstead_status_subset} and~\ref{tab:maintainability_status_subset}.

\paragraph{Compared with the GitHub Dataset.} Figures \ref{git_maintain} and \ref{git_bugs} show that the maintainability of GitHub code increases and the estimated number of bugs began to decline after 2023 Q1, especially in non-CS repositories. Moreover, a moderate decrease can be found in the Halstead difficulty of CS repositories, as shown in Figure \ref{git_diff}. The evolution of these three metrics after 2023 Q1 is consistent with the characteristics of LLM-generated code. 

\paragraph{Execution Checking.} In addition, we also manually evaluated the results of LLM generation and rewritten code (as shown in Table \ref{tab:accuracy_case}), and found that REF outperformed ANS on most models and questions (possibly passing problems that ANS failed, or passing more test sets, and generally using less time and memory, even less than AC). This also shows that our analysis of style is consistent with the performance of the code.

\finding{LLM-generated code tends to exhibit higher maintainability, lower difficulty, and fewer bugs than human-written solutions, which aligns with the evolution of Github code. }

\section{View III: Code Similarity}
\subsection{Settings}

To quantify how closely LLM outputs resemble human style and to assess the impact of conditioning on a human solution, we compare three versions of the code of each problem: the original human-authored solution (\textbf{AC}), the LLM’s output given only the problem description (\textbf{ANS}) and the LLM’s output when additionally conditioned on the human solution (\textbf{REF}). The code generation and rewriting strategies are described in Section~\ref{generation_strategies}. 

 \begin{table*}[!t]
\small
\setlength{\tabcolsep}{6pt}
\renewcommand{\arraystretch}{1.2}
\centering
\begin{tabular}{l|cc|cc|cc|cc}
\toprule
\textbf{Model} 
& \multicolumn{2}{c|}{\textbf{ANS (Python)}} 
& \multicolumn{2}{c|}{\textbf{REF (Python)}} 
& \multicolumn{2}{c|}{\textbf{ANS (C/C++)}} 
& \multicolumn{2}{c}{\textbf{REF (C/C++)}} \\
\cmidrule(lr){2-3} \cmidrule(lr){4-5} \cmidrule(lr){6-7} \cmidrule(lr){8-9}
& Match & Error & Match & Error & Match & Error & Match & Error \\
\midrule
Qwen3-32B     & 18.28\% & 25.46\% & 13.11\% & 16.44\% & 26.89\% & 33.99\% & 25.59\% & 23.76\% \\
Gemma-3-27B    & 10.68\% & 17.28\% & 12.94\% & 15.78\% & 16.03\% & 19.79\% & 23.45\% & 21.17\% \\
Deepseek-R1-Distill-Qwen-32B & 14.77\% & 20.20\% & 13.02\% & 14.86\% & 28.00\% & 30.59\% & \textbf{30.95\%} & 25.73\% \\
GPT-4.1      & \textbf{22.54\%} & \textbf{29.30\%} & \textbf{16.36\%} & \textbf{23.21\%} & \textbf{38.14\%} & \textbf{49.75\%} & 29.79\% & \textbf{27.51\%} \\
Gemini-2.0-Flash   & 12.60\% & 15.61\% & 12.27\% & 12.27\% & 21.75\% & 21.84\% & 22.73\% & 16.12\% \\
Llama-4-Maverick   & 12.10\% & 15.61\% &  9.85\% & 12.35\% & 21.08\% & 23.58\% & 23.63\% & 18.98\% \\
\bottomrule
\end{tabular}
\caption{Match and error rates (defined in Section \ref{tasks}) between the model's predicted reasoning (ANS or REF) and the ground-truth algorithm labels, for both Python and C/C++ code.}
\label{match}
\end{table*}

\subsection{Results}

Tables~\ref{cos} and~\ref{jaccard} report the cosine and Jaccard similarities among AC, ANS, and REF.  We can see that the overall trends of cosine similarity and Jaccard similarity are consistent. Among the three pairwise comparisons, AC versus REF yields the highest similarity, indicating that \textbf{LLMs are capable of imitating a given human-written solution when provided}.

In contrast, AC vs. ANS exhibits the lowest similarity and remains relatively low overall, suggesting that in the context of IO algorithm programming tasks, LLM-generated code, when produced without reference, \textbf{differs substantially in style from human-written code}. 

Additionally, similarity scores vary across models. For instance, in the AC vs. REF comparison, \textit{Llama-4-Maverick} scored the highest in both languages among all models, while for the AC vs. ANS setting, Deepseek and Gemini score the highest in C/C++ and Python, respectively. 

These results show that different LLMs have different similarities with human code and different abilities to learn from and imitate human solutions.

\finding{LLMs can effectively mimic human coding style when given reference code, but without such guidance, their generated solutions diverge significantly from human-written code in IO algorithm tasks.}

\section{Labels in the Reasoning Process} 
\subsection{Settings}

In addition to the final generated code, the reasoning process of LLMs can also reveal how LLMs understand and solve coding problems.

First, each problem on codeforces is associated with one or more algorithm labels, corresponding to different potential solution strategies. Since the original \texttt{code4bench} dataset does not contain this part, we have extended it by incorporating these labels. Then, for the LLMs' reasoning process, we check whether these labels are explicitly mentioned.
After counting each problem, we calculate the match rate and error rate of questions of various difficulty levels. 

\begin{figure}[h]
    \centering
        \includegraphics[width=0.46\textwidth]{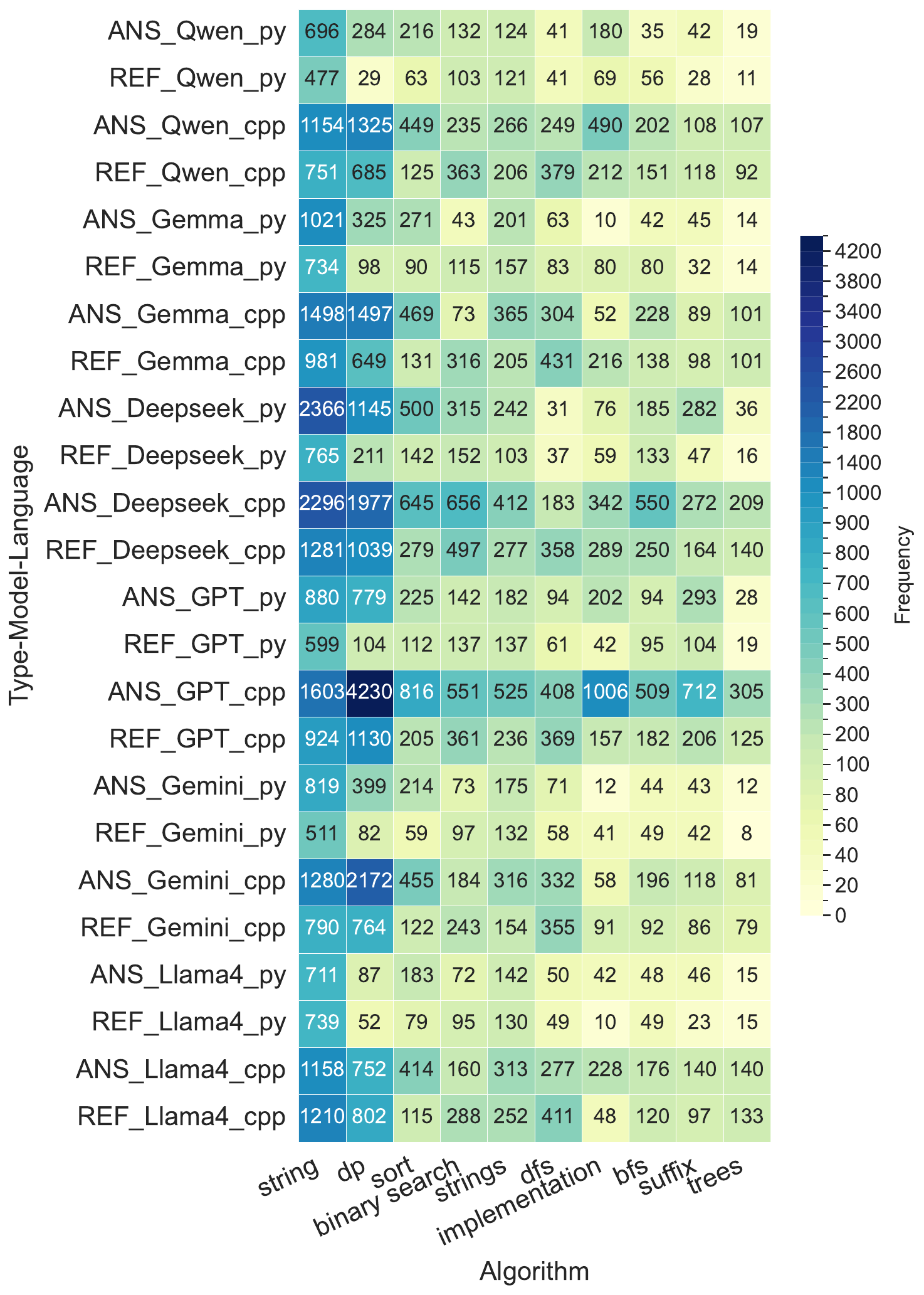}
    \caption{Frequency comparison of top 10 algorithms on various models (ANS/REF, Python/C/C++).}
    \label{label_freq_10}
\end{figure}

\subsection{Results}

\paragraph{Label Frequencies.} Figure~\ref{label_freq_10} illustrates the frequency of the 10 most common algorithm labels in different domains for each model. Table~\ref{all_tag} presents the label frequency of the collected questions, while Table~\ref{tag_frequencies} and Table~\ref{tag_frequencies_continue} show the label frequency during the LLM reasoning process. Most of the frequencies in ANS are higher than those in REF, indicating that \textbf{the reference code-based model tends to analyze without relying on algorithms}. The label frequency in C/C++ is always higher than that in Python, suggesting that the \textbf{model is accustomed to analyzing from an algorithmic perspective when implementing C/C++ code}. Furthermore, GPT's output frequency in these ten labels is higher than other models, and \textit{dp} even exceed 4000, implying that \textbf{its algorithm-based thinking performs best in IO scenarios}.

\paragraph{Match and Error Rate.}
Table~\ref{match} shows the match and error rates between the ground-truth labels and the reasoning outputs of the LLMs.  The detailed results at different levels of difficulty are presented in Table~\ref{tab:match_error_diffi}. From the results, we observe that the error rate generally exceeds the matching rate, suggesting that \textbf{LLMs tend to explore more incorrect approaches}, likely due to their reliance on a limited set of mainstream algorithms. 

Moreover, the match rate and error rate of most REF cases are lower than those of ANS, indicating that \textbf{most models tend to imitate and refine existing solutions when provided with reference code}. GPT consistently achieves the highest match rate and error rate compared to other models, meaning that it is more inclined to use algorithmic thinking to analyze IO questions. 

Furthermore, both the match and error rates are higher for C/C++ than for Python, which may reflect \textbf{language-specific design choices in the models-favoring algorithmic reasoning in C/C++ and practical implementation in Python}. Finally, as problem difficulty increases, both match and error rates rise across models and types.

\finding{LLMs have low algorithm analysis capabilities, are more inclined to approach C/C++ code from an algorithmic perspective, and harder problems may better activate their algorithmic reasoning capabilities.}

\section{Related Work}
\paragraph{LLMs for Code Generation.} Code generation has been seen rapid progress in recent years. Before ChatGPT arrived, transformer-based models for code generation had already been developed, such as CodeBERT~\citep{feng2020codebert}, CodeT5~\citep{wang2021codet5}, Codex~\citep{chen2021evaluating}, AlphaCode~\citep{li2022competition}.

\paragraph{Code Generation Evaluation.} There has also been extensive research on how to evaluate and compare the code generation capabilities of LLMs~\citep{lu2021codexglue,vaithilingam2022expectation,jimenez2023swe,dong2025codescore}. The discussion about the code generation capabilities of LLMs has never stopped~\citep{manh2024codemmlu,zheng2024makes,chen2025memorize,tambon2025bugs}, which is considered by the broader research community as a key evaluation metric for new LLMs~\citep{deepseekai2025deepseekr1incentivizingreasoningcapability,yang2025qwen3}. 

\paragraph{LLM-Generated Code Detection.} The methods for detecting the code generated by LLMs are diverse, such as feature-based classifiers~\citep{rahman2024automatic,demirok2024aigcodeset,park2025detection}, contrastive learning~\citep{ye2024uncovering,xu2024distinguishing}, transformer-based encoder classifier~\citep{gurioli2024you}. People are also interested in lexical diversity, readability, perplexity, conciseness, and naturalness~\citep{shi2024between,wang2024beyond,xu2024detecting}. 
However, many studies have pointed out the limitations of existing detectors~\citep{xu2024investigating,suh2024empirical}, encouraging us to explore the influence of LLMs on code style rather than detection.

\section{Discussion and Conclusion}
LLMs have a wide variety of applications in scientific research~\citep{eger2025transforming}. The impact of LLMs, particularly in academic writing, has been measured in various ways~\citep{liang2024monitoring,geng2024chatgpt}. The use of LLM-assisted programming tools is likely to increase, but the effect of LLMs on code, especially when using real-world data, remains to be explored.

The number of questions on Stack Overflow has declined since the emergence of LLMs\footnote{\url{https://blog.pragmaticengineer.com/stack-overflow-is-almost-dead/}}, which has raised concerns among many~\citep{zhong2024can}. LLMs could improve coding productivity~\citep{ziegler2022productivity}, and effectively support students in learning coding~\citep{korpimies2024unrestricted,rasnayaka2024empirical}, although several issues remain unresolved~\citep{pearce2025asleep}.

Our simulation findings reaffirm that LLM-generated code possesses its own features when compared to human-written code. There are both differences and similarities among various LLMs, such as in the names of variables and functions. Therefore, we attempt to identify traces of LLMs in the code from GitHub repositories. We find the first large-scale empirical evidence that LLMs affect real-world programming style, although it is impossible to determine the specific numerical value of the user percentage. 

Consequently, there is a strong possibility that human coding style will shift toward that of LLMs in the future. We therefore emphasize the need to consider not only the programming capabilities of LLMs, but also their broader societal implications. 

\section*{Limitations}

Although we have conducted multiple experiments to evaluate the changes in code style in the LLM era, our research still has some limitations. 

First, We did not examine how much of the code on GitHub was generated by LLMs, as this is not the focus of our study. While such an analysis could potentially provide additional support for our findings, there is currently a lack of widely adopted methods and benchmarks.

Secondly, our dataset can be further expanded. Code4bench is an early user AC code collection, which reflects the poor code style and thinking ability of users in the recent LLM era. The code collection of a single evaluation result cannot reflect the overall style of the entire human code ecology, and the comparison with the LLM-generated code is not complete. We will rebuild our code data set.

In addition, there may be many scenarios for user-generated code and prompt parameters. Our simulation cannot exhaust all user usage situations, and the universality of our research results needs to be further improved.

\section*{Ethical Considerations}
Our focus is on examining the impact of LLMs on domain-specific coding styles, rather than identifying whether a given piece of code was generated by LLMs. This is not only due to technical considerations, but also to reduce potential ethical risks.

\bibliography{custom}

\appendix

\section{Dataset}

\paragraph{GitHub Data} There are some issues in our dataset collection process. We use regular expressions to detect whether the abstract of each paper in the arXiv dataset contains a GitHub repository link. In the arXiv dataset\footnote{The links to these repositories are mentioned in the metadata of the arXiv dataset: \url{https://www.kaggle.com/datasets/Cornell-University/arxiv/data}}, a single GitHub link may appear multiple times, likely because the same repository was used for multiple paper submissions. For such repositories, if the publication dates of all associated papers fall within a two-quarter range, we retain the link and assign it the most recent publication date as its timestamp. Otherwise, we discard the repository. Repositories that lack target language files or are excessively large are also excluded from our analysis.

\begin{figure*}[ht]
    \centering
\begin{tcolorbox}[enhanced,attach boxed title to top center={yshift=-3mm,yshifttext=-1mm},boxrule=0.9pt, title=SQL Queries,
  colback=gray!00,colframe=black!50,colbacktitle=gray,
  boxed title style={size=small,colframe=gray} ]

SELECT DISTINCT s.id, s.sourceCode, s.languages\_id, p.fullname, p.id, p.name, s.countline, p.context

FROM code4bench.source s

JOIN code4bench.problems p ON s.problems\_id = p.id

WHERE s.languages\_id IN (1, 8)
  
  AND s.verdicts\_id = (
    
    SELECT id
    
    FROM code4bench.verdicts
    
    WHERE name = 'Accepted'
  
  )
  
  AND s.isduplicated = 0;

\end{tcolorbox}
    \caption{SQL queries for retrieving accepted Python and C/C++ code submissions from the Code4Bench database.}
    \label{SQL}
\end{figure*}

\begin{figure}[t]
    \centering
    \includegraphics[width=\linewidth]{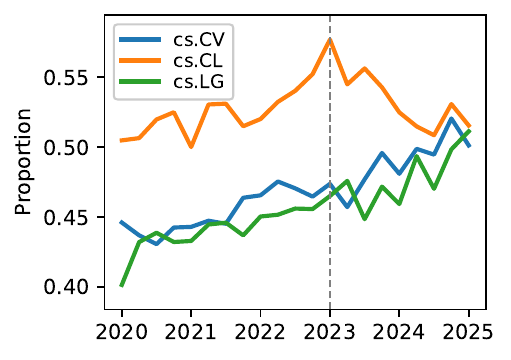}
    \caption{\textit{snake\_case} variable across different CS repos.}
    \label{detail_var}
\end{figure}

\begin{figure}[t]
    \centering
    \includegraphics[width=\linewidth]{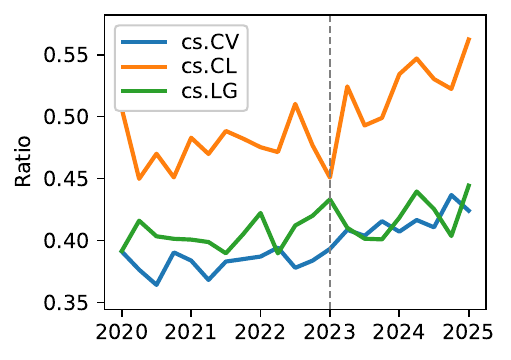}
    \caption{\textit{snake\_case} function across different CS repos.}
    \label{detail_func}
\end{figure}

Table~\ref{tab:quarter_repo_stats} shows the number of repositories in our dataset. Table~\ref{subset} shows the number of problems per main algorithm across difficulty buckets.

\paragraph{Human-Written Solutions} We retrieved accepted Python and C/C++ code submissions from the Code4Bench database using SQL queries. The SQL is shown in Figure~\ref{SQL}. For each Codeforces problem, we first retain only a single human-written submission. We then use \textit{BeautifulSoup} to convert the problem description from HTML into plain text.

\begin{table*}[!t]
\small
\renewcommand{\arraystretch}{1.1}
\centering
\begin{tabular}{l|cccc}
\toprule
\textbf{Quarter} & \textbf{Python-cs} & \textbf{Python-non-cs} & \textbf{C/C++-cs} & \textbf{C/C++-non-cs} \\
& (\#Repo / \#Files) & (\#Repo / \#Files) & (\#Repo / \#Files) & (\#Repo / \#Files) \\
\midrule
2020 Q1 & 462 / 21643 & 139 / 3998 & 81 / 1532 & 29 / 385 \\
2020 Q2 & 488 / 27190 & 111 / 2755 & 85 / 1002 & 18 / 241 \\
2020 Q3 & 468 / 23008 & 152 / 5188 & 78 / 1077 & 18 / 206 \\
2020 Q4 & 499 / 35083 & 131 / 3857 & 91 / 1367 & 19 / 490 \\
2021 Q1 & 480 / 27155 & 139 / 3025 & 93 / 962 & 16 / 324 \\
2021 Q2 & 523 / 28696 & 135 / 3471 & 117 / 1332 & 21 / 295 \\
2021 Q3 & 520 / 27802 & 127 / 2448 & 121 / 1544 & 20 / 139 \\
2021 Q4 & 508 / 29693 & 160 / 3649 & 105 / 1247 & 26 / 357 \\
2022 Q1 & 486 / 24890 & 185 / 5684 & 160 / 1612 & 29 / 456 \\
2022 Q2 & 498 / 31272 & 192 / 5495 & 170 / 2450 & 35 / 599 \\
2022 Q3 & 495 / 34820 & 202 / 6358 & 208 / 3233 & 29 / 308 \\
2022 Q4 & 515 / 37587 & 193 / 4873 & 218 / 2354 & 18 / 341 \\
2023 Q1 & 521 / 45587 & 191 / 5862 & 225 / 2150 & 24 / 185 \\
2023 Q2 & 508 / 38882 & 240 / 6345 & 226 / 2416 & 35 / 872 \\
2023 Q3 & 506 / 36300 & 255 / 6110 & 267 / 2429 & 25 / 186 \\
2023 Q4 & 525 / 44676 & 232 / 7428 & 277 / 3177 & 31 / 212 \\
2024 Q1 & 524 / 34320 & 250 / 6581 & 266 / 3058 & 24 / 301 \\
2024 Q2 & 533 / 44075 & 325 / 9628 & 320 / 4070 & 41 / 616 \\
2024 Q3 & 530 / 48355 & 359 / 9620 & 366 / 5541 & 49 / 542 \\
2024 Q4 & 529 / 40706 & 431 / 13031 & 405 / 4363 & 50 / 957 \\
2025 Q1 & 520 / 59170 & 323 / 10541 & 296 / 4363 & 56 / 787 \\
2025 Q2 & 777 / 45674 & 116 / 5868 & 139 / 2191 & 21 / 852 \\
2025 Q3 & 752 / 41810 & 157 / 6868 & 173 / 1340 & 41 / 478 \\
\bottomrule
\end{tabular}
\caption{Number of repositories and Python/C++ files per quarter and category}
\label{tab:quarter_repo_stats}
\end{table*}

\begin{table*}[!t]
\centering
\renewcommand{\arraystretch}{1.1}
\begin{tabular}{@{}lcccc@{}}
\toprule[1.5pt]
\textbf{Main Algorithm} & \textbf{800--1199} & \textbf{1200--1599} & \textbf{1600--1999} & \textbf{2000+} \\
\midrule
implementation             & 25 & 13 & 6  & 2  \\
brute force                & 10 & 9  & 7  & 12 \\
constructive algorithms    & 5  & 6  & 9  & 7  \\
greedy                     & 5  & 8  & 4  & 1  \\
math                       & 3  & 2  & 3  & 4  \\
binary search              & 1  & 4  & 8  & 9  \\
dp                         & 1  & 2  & 4  & 7  \\
data structures            & 0  & 2  & 4  & 3  \\
combinatorics              & 0  & 2  & 2  & 4  \\
dfs and similar            & 0  & 2  & 3  & 1  \\
\bottomrule[1.5pt]
\end{tabular}
\caption{Number of problems per main algorithm across difficulty buckets.}
\label{subset}
\end{table*}

\subsection{Prompt}

Figures \ref{prompt_generate_1}, \ref{prompt_revise}, and \ref{prompt_generate_2} present the prompt designs for direct generation, reference-guided generation, and large-scale generation, respectively.

\begin{figure*}[!t]
    \centering
\begin{tcolorbox}[enhanced,attach boxed title to top center={yshift=-3mm,yshifttext=-1mm},boxrule=0.9pt, title=Prompt 1: Direct Generation,
  colback=gray!00,colframe=black!50,colbacktitle=gray,
  boxed title style={size=small,colframe=gray} ]

Your task is to carefully read the following problem description and implement a solution in \{language\}.\\

Please first provide your reasoning in plain text, and then provide the corresponding code.\\

Format your response as follows using Markdown:\\

\#\#\# Reasoning

<Please provide only your step-by-step reasoning in plain text here.>\\

\#\#\# Code

<Please provide only your code in \{language\} here, with no extra explanation or text.>\\

Here is the problem description:\\
\{context\}

\end{tcolorbox}
    \caption{Prompt instructing LLMs to provide reasoning and solution code based only on problem descriptions.}
    \label{prompt_generate_1}
\end{figure*}

\begin{figure*}[!t]
    \centering
\begin{tcolorbox}[enhanced,attach boxed title to top center={yshift=-3mm,yshifttext=-1mm},boxrule=0.9pt, title=Prompt 2: Reference-Guided Generation,
  colback=gray!00,colframe=black!50,colbacktitle=gray,
  boxed title style={size=small,colframe=gray} ]

Your task is to carefully read the following problem description and revise the given code. \\

The code given is AC code (correct and has passed the test.) \\

Please first provide your reasoning in plain text, and then provide the corresponding code. \\

Format your response as follows using Markdown:\\

\#\#\# Reasoning

<Please provide only your step-by-step reasoning in plain text here.>\\

\#\#\# Code

<Please provide only your code in \{language\} here, with no extra explanation or text.>\\

Here is the problem description:

\{context\}\\

Here is the user's AC Code:

\{code\}

\end{tcolorbox}
    \caption{Prompt instructing LLMs to provide reasoning process and solution code based on both problem descriptions and correct human-written code.}
    \label{prompt_revise}
\end{figure*}

\begin{figure*}[!t]
    \centering
\begin{tcolorbox}[enhanced,attach boxed title to top center={yshift=-3mm,yshifttext=-1mm},boxrule=0.9pt, title=Prompt 3: Large-Scale Generation,
  colback=gray!00,colframe=black!50,colbacktitle=gray,
  boxed title style={size=small,colframe=gray} ]

Your task is to carefully read the following problem description and implement a solution in \{language\}. Return only the code without any explanations. \\

Here is the problem description:\\
\{context\}

\end{tcolorbox}
    \caption{Prompt instructing LLMs to provide only solution code based on problem descriptions.}
    \label{prompt_generate_2}
\end{figure*}

\section{Github Result}
\label{github naming}

\begin{figure*}[!t]
  \centering
  \begin{subfigure}[t]{0.49\linewidth}
    \centering
    \begin{subfigure}[t]{0.49\linewidth}
      \centering
      \includegraphics[width=\linewidth]{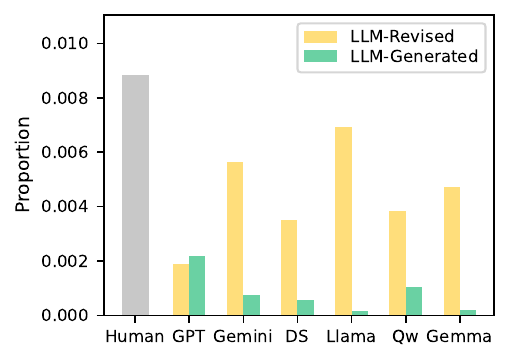}
    \end{subfigure}\hfill
    \begin{subfigure}[t]{0.49\linewidth}
      \centering
      \includegraphics[width=\linewidth]{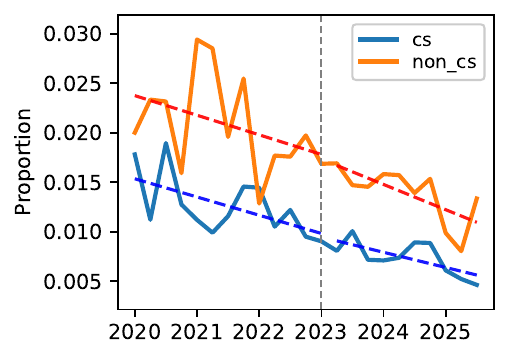}
    \end{subfigure}
    \caption{camelCase variables.}
  \end{subfigure}
  \hfill
  \begin{subfigure}[t]{0.49\linewidth}
    \centering
    \begin{subfigure}[t]{0.49\linewidth}
      \centering
      \includegraphics[width=\linewidth]{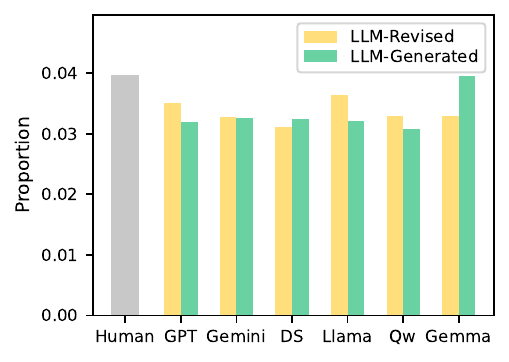}
    \end{subfigure}\hfill
    \begin{subfigure}[t]{0.49\linewidth}
      \centering
      \includegraphics[width=\linewidth]{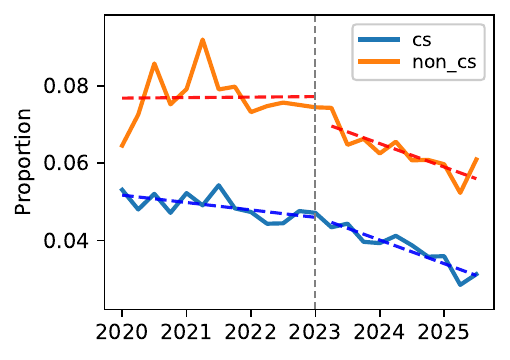}
    \end{subfigure}
    \caption{Digit-suffixed variables.}
    \label{python_var_digit}
  \end{subfigure}

  \begin{subfigure}[t]{0.49\linewidth}
    \centering
    \begin{subfigure}[t]{0.49\linewidth}
      \centering
      \includegraphics[width=\linewidth]{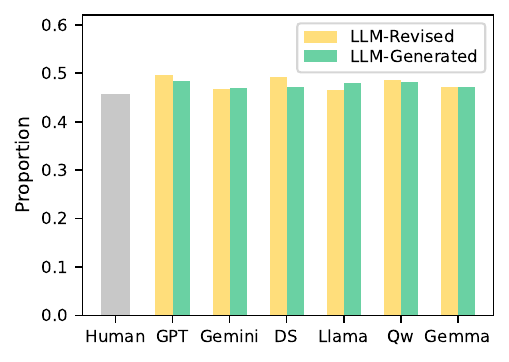}
    \end{subfigure}\hfill
    \begin{subfigure}[t]{0.49\linewidth}
      \centering
      \includegraphics[width=\linewidth]{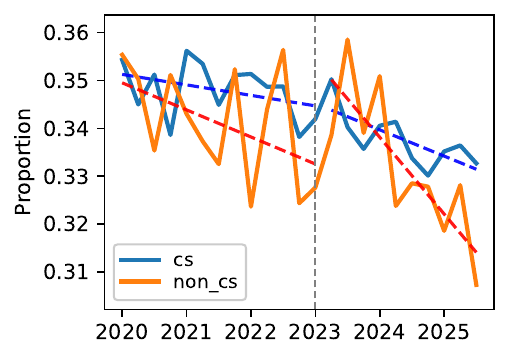}
    \end{subfigure}
    \caption{Lowercase variables.}
  \end{subfigure}
  \hfill
  \begin{subfigure}[t]{0.49\linewidth}
    \centering
    \begin{subfigure}[t]{0.49\linewidth}
      \centering
      \includegraphics[width=\linewidth]{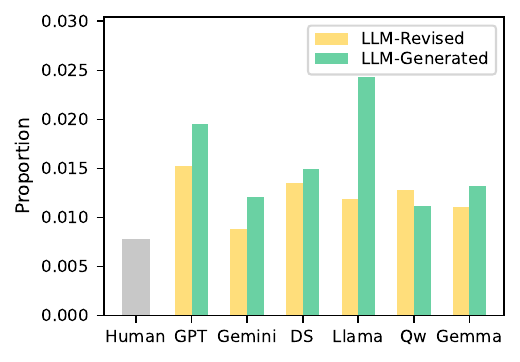}
    \end{subfigure}\hfill
    \begin{subfigure}[t]{0.49\linewidth}
      \centering
      \includegraphics[width=\linewidth]{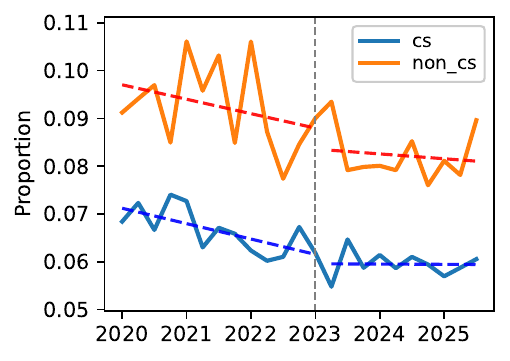}
    \end{subfigure}
    \caption{Other variables.}
  \end{subfigure}

  \begin{subfigure}[t]{0.49\linewidth}
    \centering
    \begin{subfigure}[t]{0.49\linewidth}
      \centering
      \includegraphics[width=\linewidth]{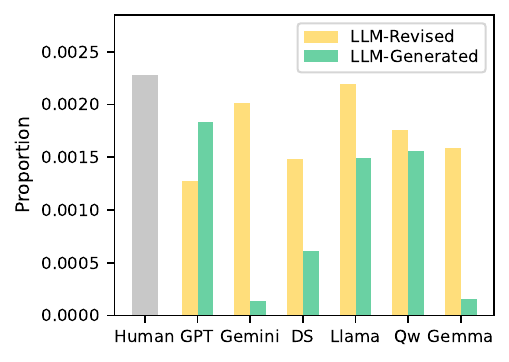}
    \end{subfigure}\hfill
    \begin{subfigure}[t]{0.49\linewidth}
      \centering
      \includegraphics[width=\linewidth]{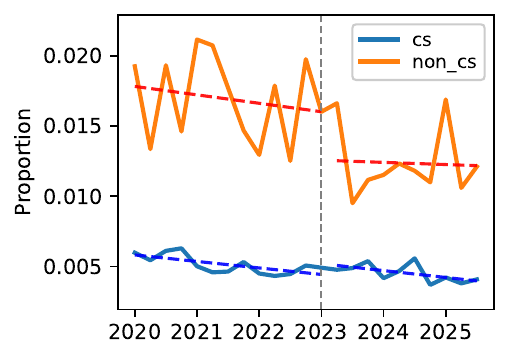}
    \end{subfigure}
    \caption{PascalCase variables.}
  \end{subfigure}
  \hfill
  \begin{subfigure}[t]{0.49\linewidth}
    \centering
    \begin{subfigure}[t]{0.49\linewidth}
      \centering
      \includegraphics[width=\linewidth]{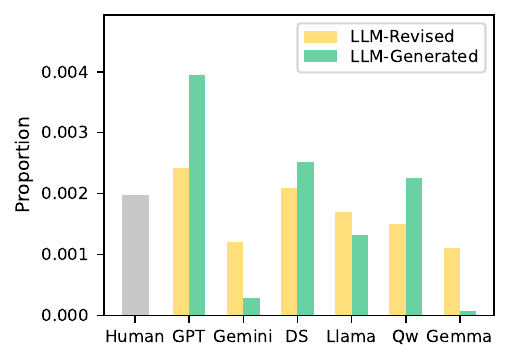}
    \end{subfigure}\hfill
    \begin{subfigure}[t]{0.49\linewidth}
      \centering
      \includegraphics[width=\linewidth]{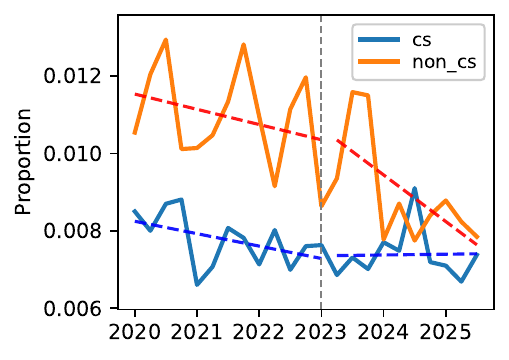}
    \end{subfigure}
    \caption{UPPERCASE variables.}
  \end{subfigure}
  \begin{subfigure}[t]{0.49\linewidth}
    \centering
    \begin{subfigure}[t]{0.49\linewidth}
      \centering
      \includegraphics[width=\linewidth]{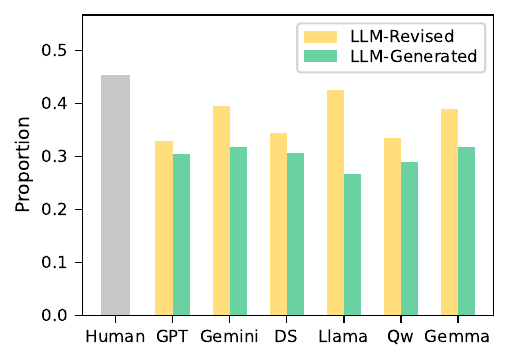}
    \end{subfigure}\hfill
    \begin{subfigure}[t]{0.49\linewidth}
      \centering
      \includegraphics[width=\linewidth]{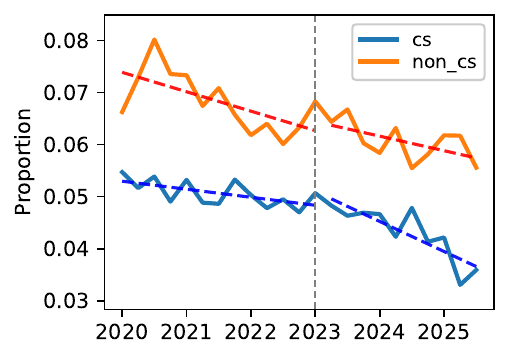}
    \end{subfigure}
    \caption{Single letter variables.}
    \label{python_var_single}
  \end{subfigure}\hfill
\hfill
  \begin{subfigure}[t]{0.49\linewidth}
    \centering
    \begin{subfigure}[t]{0.49\linewidth}
      \centering
      \includegraphics[width=\linewidth]{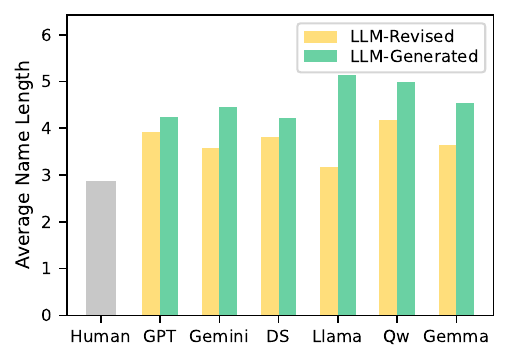}
    \end{subfigure}\hfill
    \begin{subfigure}[t]{0.49\linewidth}
      \centering
      \includegraphics[width=\linewidth]{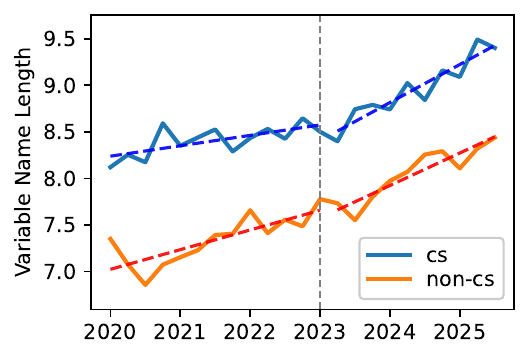}
    \end{subfigure}
    \caption{Average length.}
    \label{py_var_len}
  \end{subfigure}\hfill

  \caption{Comparison of \textbf{Python variable} naming styles in LLM‐generated vs.\ human‐written code and their temporal trends on GitHub. Notably, LLMs exhibit \textbf{a general avoidance of camelCase, digit-suffixed, and single-letter variables}, with a marked decline in usage frequency on GitHub repositories. The consistent trends across both datasets suggest a potential correlation between LLM preferences and real-world coding practices.}
  \label{Python variable}
\end{figure*}

\begin{figure*}[!t]
    \centering

    \begin{subfigure}[t]{0.49\linewidth}
        \centering
        \begin{subfigure}[t]{0.49\linewidth}
            \centering
            \includegraphics[width=\linewidth]{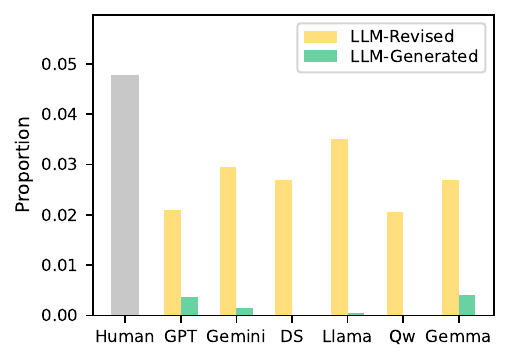}
        \end{subfigure}\hfill
        \begin{subfigure}[t]{0.49\linewidth}
            \centering
            \includegraphics[width=\linewidth]{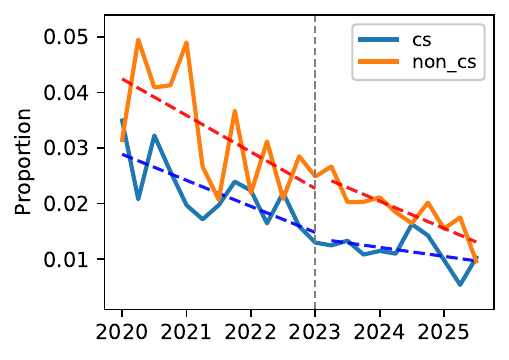}
        \end{subfigure}
        \caption{camelCase functions.}
    \end{subfigure}\hfill
    \begin{subfigure}[t]{0.49\linewidth}
        \centering
        \begin{subfigure}[t]{0.49\linewidth}
            \centering
            \includegraphics[width=\linewidth]{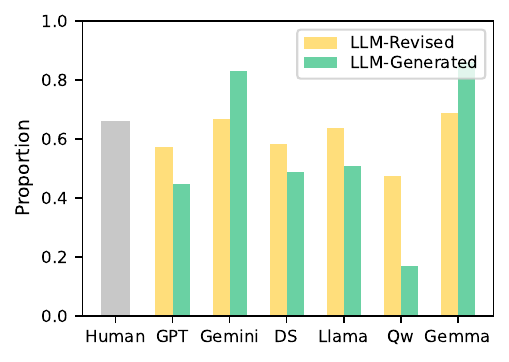}
        \end{subfigure}\hfill
        \begin{subfigure}[t]{0.49\linewidth}
            \centering
            \includegraphics[width=\linewidth]{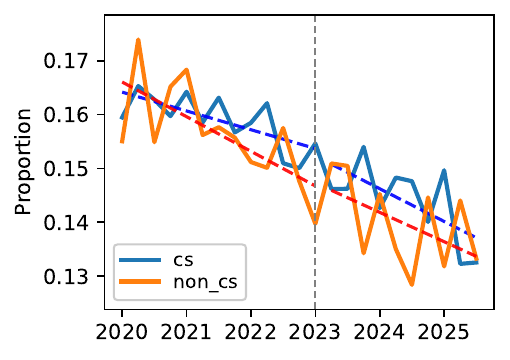}
        \end{subfigure}
        \caption{Lowercase functions.}
    \end{subfigure}\hfill

    \begin{subfigure}[t]{0.49\linewidth}
        \centering
        \begin{subfigure}[t]{0.49\linewidth}
            \centering
            \includegraphics[width=\linewidth]{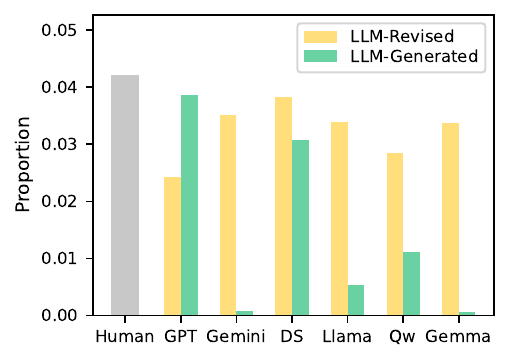}
        \end{subfigure}\hfill
        \begin{subfigure}[t]{0.49\linewidth}
            \centering
            \includegraphics[width=\linewidth]{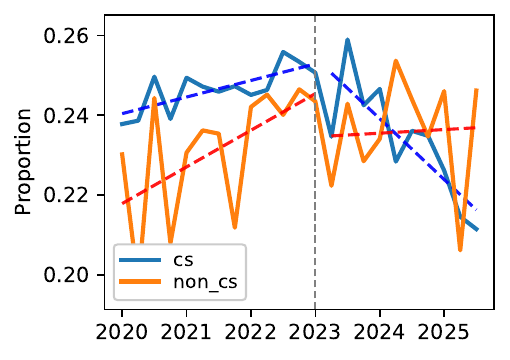}
        \end{subfigure}
        \caption{Other functions.}
    \end{subfigure}
\hfill
    \begin{subfigure}[t]{0.49\linewidth}
        \centering
        \begin{subfigure}[t]{0.49\linewidth}
            \centering
            \includegraphics[width=\linewidth]{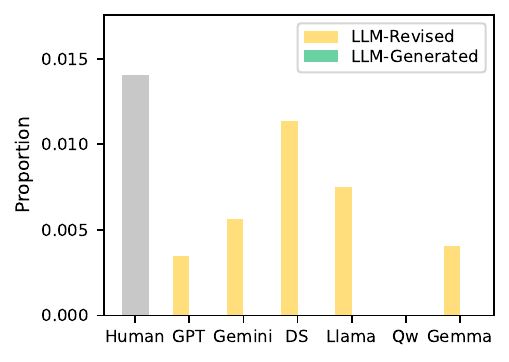}
        \end{subfigure}\hfill
        \begin{subfigure}[t]{0.49\linewidth}
            \centering
            \includegraphics[width=\linewidth]{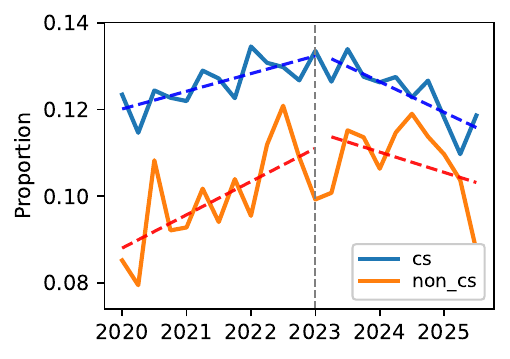}
        \end{subfigure}
        \caption{PascalCase functions.}
    \end{subfigure}

    \begin{subfigure}[t]{0.49\linewidth}
        \centering
        \begin{subfigure}[t]{0.49\linewidth}
            \centering
            \includegraphics[width=\linewidth]{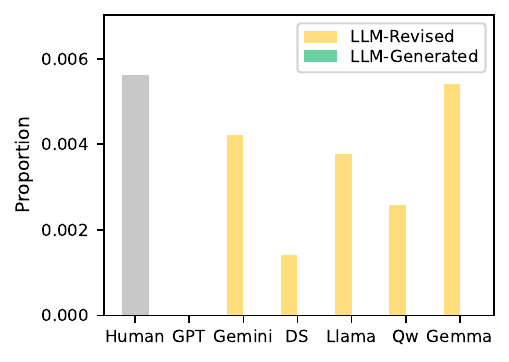}
        \end{subfigure}\hfill
        \begin{subfigure}[t]{0.49\linewidth}
            \centering
            \includegraphics[width=\linewidth]{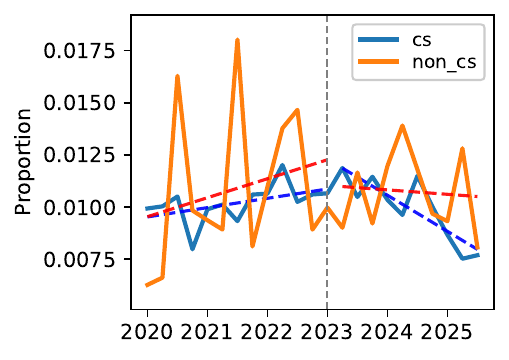}
        \end{subfigure}
        \caption{UPPERCASE functions.}
    \end{subfigure}
\hfill
    \begin{subfigure}[t]{0.49\linewidth}
        \centering
        \begin{subfigure}[t]{0.49\linewidth}
            \centering
            \includegraphics[width=\linewidth]{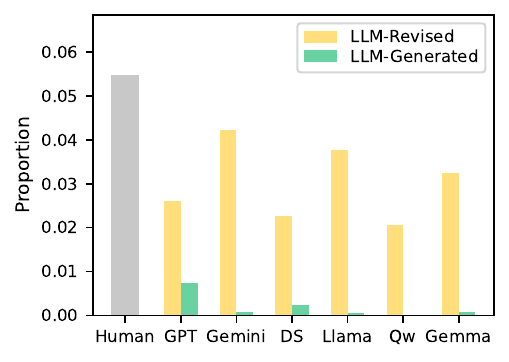}
        \end{subfigure}\hfill
        \begin{subfigure}[t]{0.49\linewidth}
            \centering
            \includegraphics[width=\linewidth]{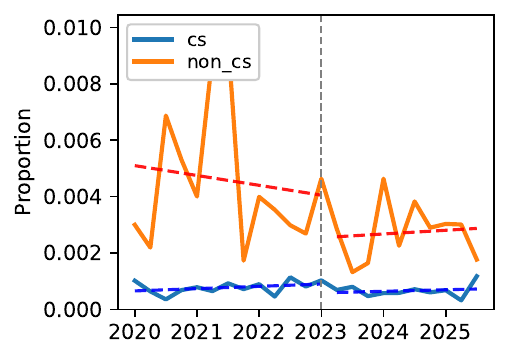}
        \end{subfigure}
        \caption{Single‐letter functions.}
    \end{subfigure}\hfill
    \begin{subfigure}[t]{0.49\linewidth}
    \end{subfigure}

    \caption{Comparison of \textbf{Python function} naming styles in LLM-generated vs.\ human-written code and their temporal trends on GitHub. This reveals that \textbf{LLMs consistently avoid the usage of camelCase, digit-suffixed, lowercase-only function names}, which is a trend paralleled by declining adoption in GitHub repositories. Furthermore, \textbf{LLMs demonstrate a clear preference for longer function names}, consistent with the increasing name length trend in GitHub repositories.}
    \label{Python function}
\end{figure*}

\begin{figure*}[!t]
    \centering
    \begin{subfigure}[t]{0.24\linewidth}
        \centering
        \includegraphics[width=\linewidth]{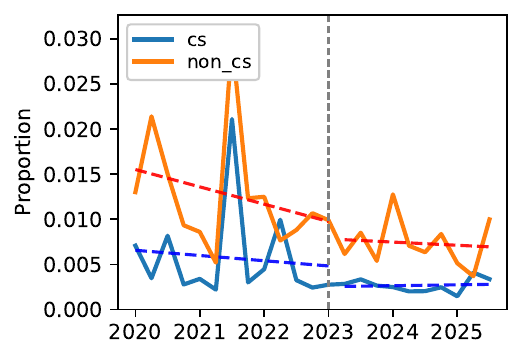}
        \caption{camelCase file names.}
    \end{subfigure}
    \hfill
    \begin{subfigure}[t]{0.24\linewidth}
        \centering
        \includegraphics[width=\linewidth]{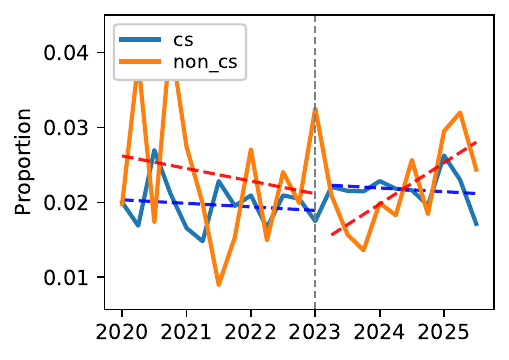}
        \caption{Digit-suffixed file names.}
    \end{subfigure}
    \hfill
    \begin{subfigure}[t]{0.24\linewidth}
        \centering
        \includegraphics[width=\linewidth]{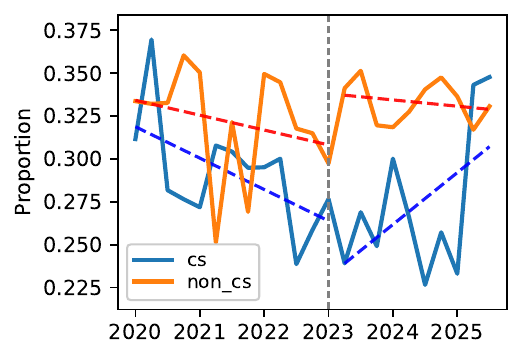}
        \caption{Lowercase file names.}
    \end{subfigure}
    \hfill
    \begin{subfigure}[t]{0.24\linewidth}
        \centering
        \includegraphics[width=\linewidth]{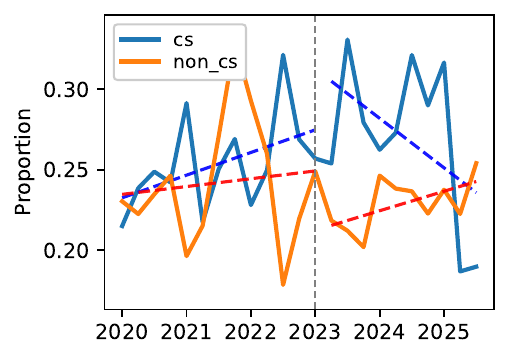}
        \caption{Other file names.}
    \end{subfigure}

    \begin{subfigure}[t]{0.24\linewidth}
        \centering
        \includegraphics[width=\linewidth]{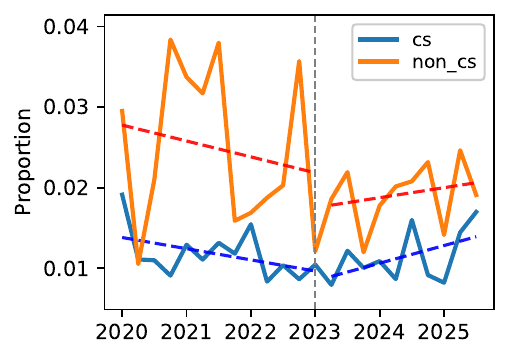}
        \caption{PascalCase file names.}
    \end{subfigure}
    \hfill
    \begin{subfigure}[t]{0.24\linewidth}
        \centering
        \includegraphics[width=\linewidth]{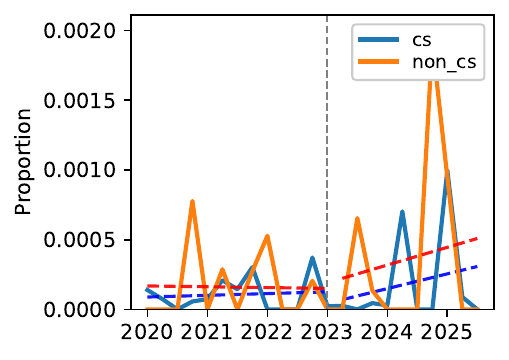}
        \caption{Single letter file names.}
    \end{subfigure}
    \hfill
    \begin{subfigure}[t]{0.24\linewidth}
        \centering
        \includegraphics[width=\linewidth]{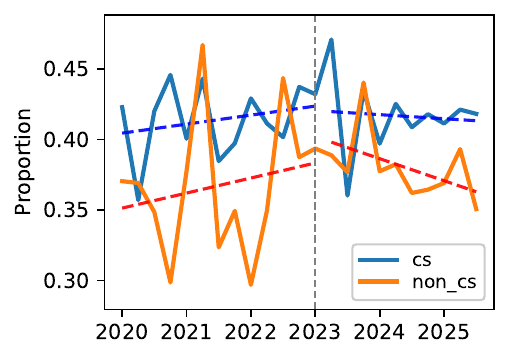}
        \caption{Snake\_case file names.}
    \end{subfigure}
    \hfill
    \begin{subfigure}[t]{0.24\linewidth}
        \centering
        \includegraphics[width=\linewidth]{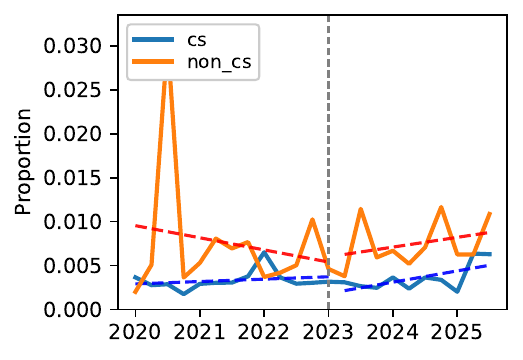}
        \caption{UPPERCASE file names.}
    \end{subfigure}  
    
    \caption{Comparison of \textbf{Python file} naming styles in LLM-generated vs. human-written code and their temporal trends on GitHub. Lowercase file names show slight decline in CS repositories, \textbf{while other naming patterns demonstrate fluctuating usage without clear trends.}}
    \label{Python file}
\end{figure*}

\begin{figure*}[!t]
  \centering
  \begin{subfigure}[t]{0.49\linewidth}
    \centering
    \begin{subfigure}[t]{0.49\linewidth}
      \centering
      \includegraphics[width=\linewidth]{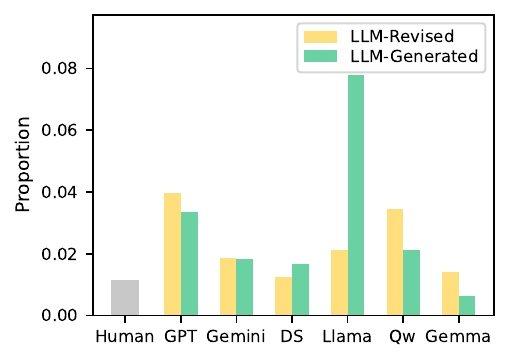}
    \end{subfigure}\hfill
    \begin{subfigure}[t]{0.49\linewidth}
      \centering
      \includegraphics[width=\linewidth]{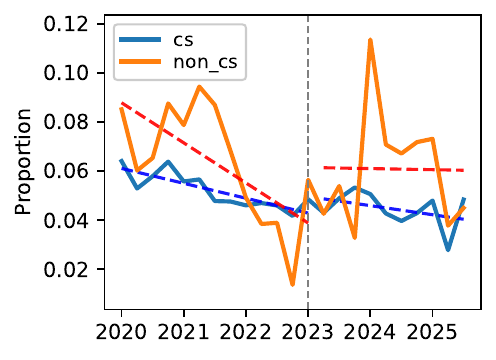}
    \end{subfigure}
    \caption{camelCase variables.}
  \end{subfigure}
  \hfill
  \begin{subfigure}[t]{0.49\linewidth}
    \centering
    \begin{subfigure}[t]{0.49\linewidth}
      \centering
      \includegraphics[width=\linewidth]{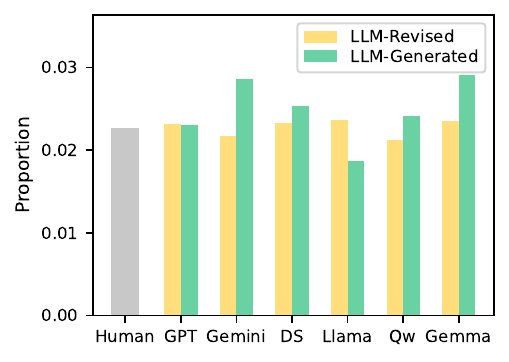}
    \end{subfigure}\hfill
    \begin{subfigure}[t]{0.49\linewidth}
      \centering
      \includegraphics[width=\linewidth]{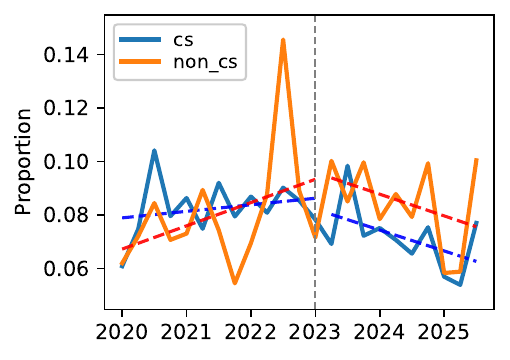}
    \end{subfigure}
    \caption{Digit-suffixed variables.}
  \end{subfigure}

  \begin{subfigure}[t]{0.49\linewidth}
    \centering
    \begin{subfigure}[t]{0.49\linewidth}
      \centering
      \includegraphics[width=\linewidth]{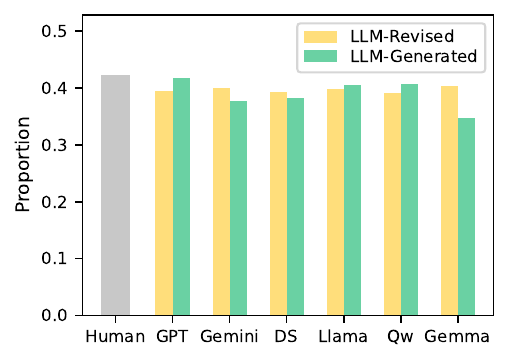}
    \end{subfigure}\hfill
    \begin{subfigure}[t]{0.49\linewidth}
      \centering
      \includegraphics[width=\linewidth]{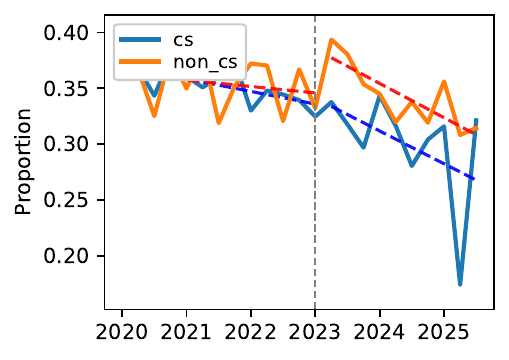}
    \end{subfigure}
    \caption{Lowercase variables.}
  \end{subfigure}
  \hfill
  \begin{subfigure}[t]{0.49\linewidth}
    \centering
    \begin{subfigure}[t]{0.49\linewidth}
      \centering
      \includegraphics[width=\linewidth]{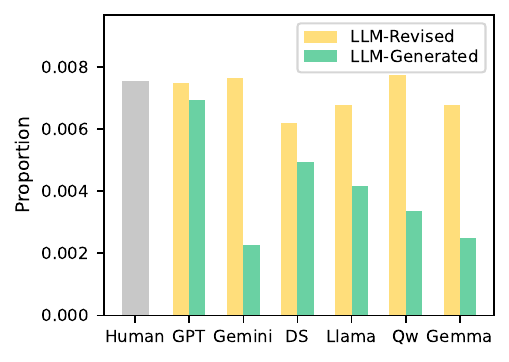}
    \end{subfigure}\hfill
    \begin{subfigure}[t]{0.49\linewidth}
      \centering
      \includegraphics[width=\linewidth]{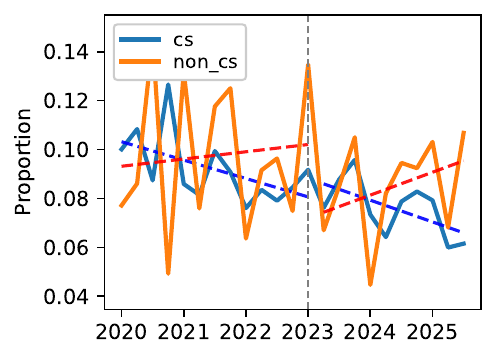}
    \end{subfigure}
    \caption{Other variables.}
  \end{subfigure}

  \begin{subfigure}[t]{0.49\linewidth}
    \centering
    \begin{subfigure}[t]{0.49\linewidth}
      \centering
      \includegraphics[width=\linewidth]{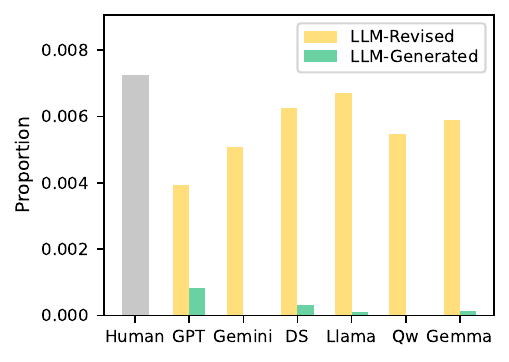}
    \end{subfigure}\hfill
    \begin{subfigure}[t]{0.49\linewidth}
      \centering
      \includegraphics[width=\linewidth]{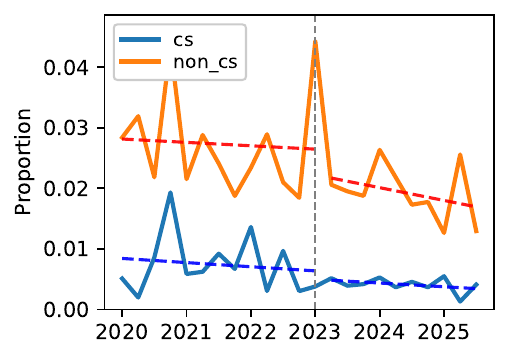}
    \end{subfigure}
    \caption{PascalCase variables.}
  \end{subfigure}
  \hfill
  \begin{subfigure}[t]{0.49\linewidth}
    \centering
    \begin{subfigure}[t]{0.49\linewidth}
      \centering
      \includegraphics[width=\linewidth]{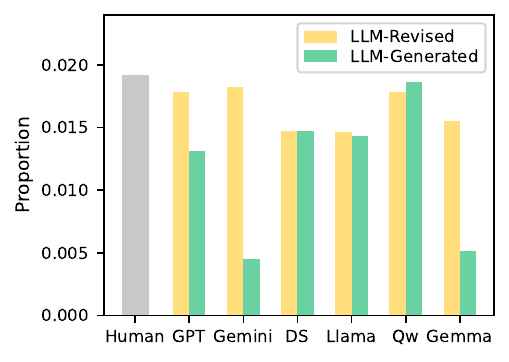}
    \end{subfigure}\hfill
    \begin{subfigure}[t]{0.49\linewidth}
      \centering
      \includegraphics[width=\linewidth]{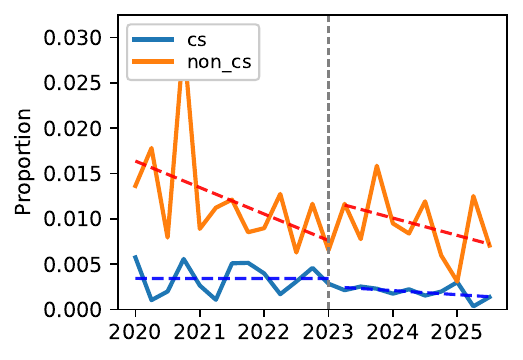}
    \end{subfigure}
    \caption{UPPERCASE variables.}
  \end{subfigure}

  \begin{subfigure}[t]{0.49\linewidth}
    \centering
    \begin{subfigure}[t]{0.49\linewidth}
      \centering
      \includegraphics[width=\linewidth]{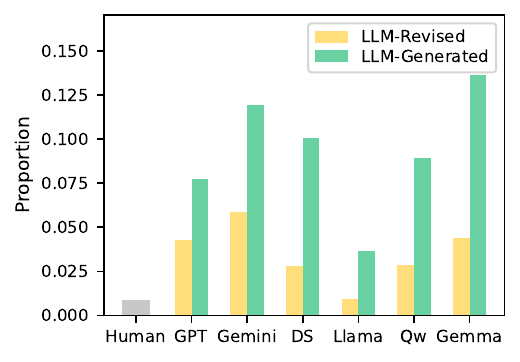}
    \end{subfigure}\hfill
    \begin{subfigure}[t]{0.49\linewidth}
      \centering
      \includegraphics[width=\linewidth]{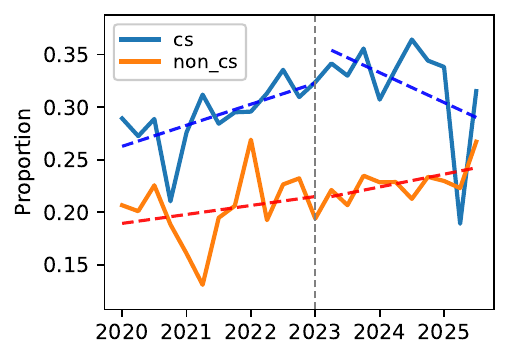}
    \end{subfigure}
    \caption{Snake\_case variables.}
    \label{cpp_var_snake}
  \end{subfigure}
  \hfill
  \begin{subfigure}[t]{0.49\linewidth}
    \centering
    \begin{subfigure}[t]{0.49\linewidth}
      \centering
      \includegraphics[width=\linewidth]{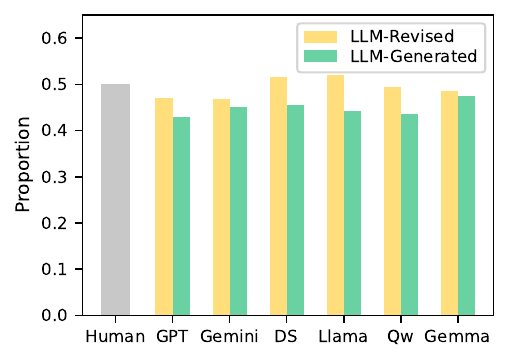}
    \end{subfigure}\hfill
    \begin{subfigure}[t]{0.49\linewidth}
      \centering
      \includegraphics[width=\linewidth]{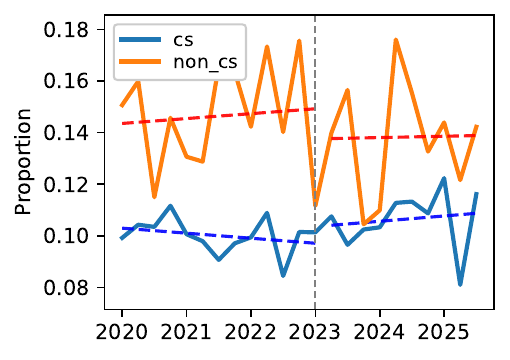}
    \end{subfigure}
    \caption{Single letter variables.}
  \end{subfigure}
  \begin{subfigure}[t]{0.49\linewidth}
    \centering
    \begin{subfigure}[t]{0.49\linewidth}
      \centering
      \includegraphics[width=\linewidth]{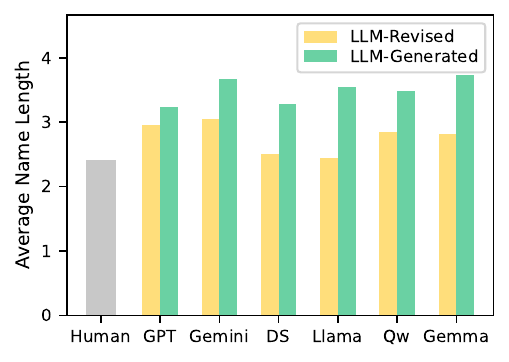}
    \end{subfigure}\hfill
    \begin{subfigure}[t]{0.49\linewidth}
      \centering
      \includegraphics[width=\linewidth]{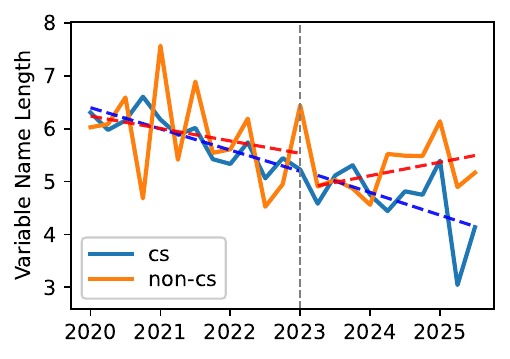}
    \end{subfigure}
    \caption{Average length.}
  \end{subfigure}\hfill
  \begin{subfigure}[t]{0.49\linewidth}
\end{subfigure}

  \caption{Comparison of \textbf{C/C++ variable} naming styles in LLM‐generated vs.\ human‐written code and their temporal trends on GitHub. \textbf{LLMs demonstrate a clear preference for snake\_case variables in C/C++.} This naming convention has shown substantial growth in GitHub repositories over the past five years. \textbf{However, we observe a gradual decline in average variable name length within these repositories}, which contrasts directly with the naming preferences exhibited by all LLM models included in our experiments.}
  \label{C/C++ variable}
\end{figure*}

\begin{figure*}[!t]
    \centering

    \begin{subfigure}[t]{0.49\linewidth}
        \centering
        \begin{subfigure}[t]{0.49\linewidth}
            \centering
            \includegraphics[width=\linewidth]{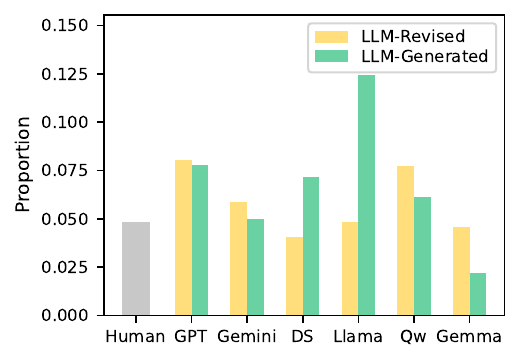}
        \end{subfigure}\hfill
        \begin{subfigure}[t]{0.49\linewidth}
            \centering
            \includegraphics[width=\linewidth]{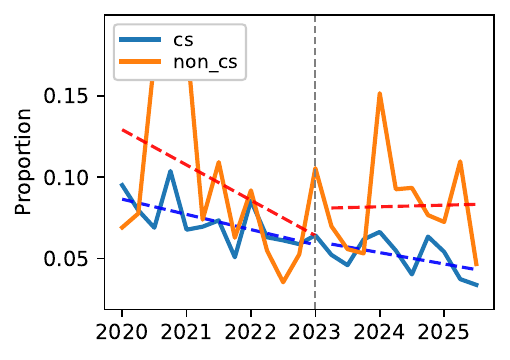}
        \end{subfigure}
        \caption{camelCase functions.}
    \end{subfigure}\hfill
    \begin{subfigure}[t]{0.49\linewidth}
        \centering
        \begin{subfigure}[t]{0.49\linewidth}
            \centering
            \includegraphics[width=\linewidth]{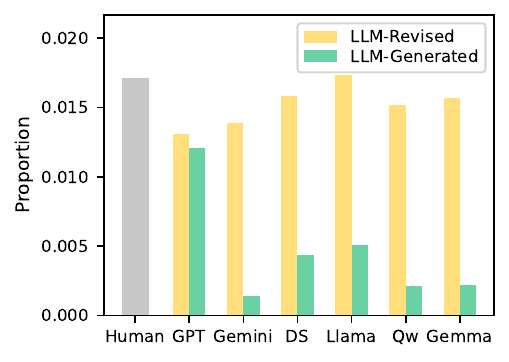}
        \end{subfigure}\hfill
        \begin{subfigure}[t]{0.49\linewidth}
            \centering
            \includegraphics[width=\linewidth]{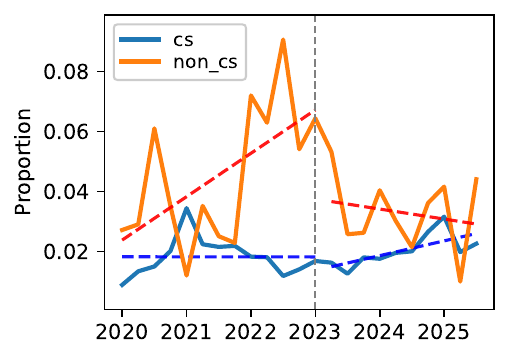}
        \end{subfigure}
        \caption{Digit‐suffixed functions.}
    \end{subfigure}

    \begin{subfigure}[t]{0.49\linewidth}
        \centering
        \begin{subfigure}[t]{0.49\linewidth}
            \centering
            \includegraphics[width=\linewidth]{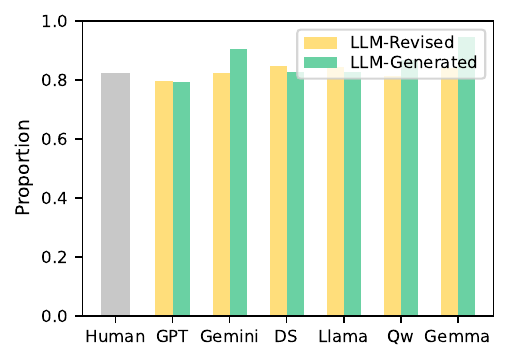}
        \end{subfigure}\hfill
        \begin{subfigure}[t]{0.49\linewidth}
            \centering
            \includegraphics[width=\linewidth]{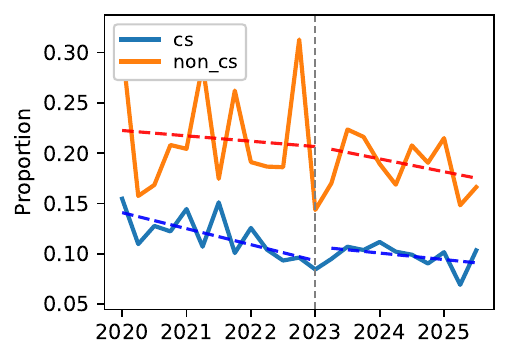}
        \end{subfigure}
        \caption{Lowercase functions.}
    \end{subfigure}\hfill
    \begin{subfigure}[t]{0.49\linewidth}
        \centering
        \begin{subfigure}[t]{0.49\linewidth}
            \centering
            \includegraphics[width=\linewidth]{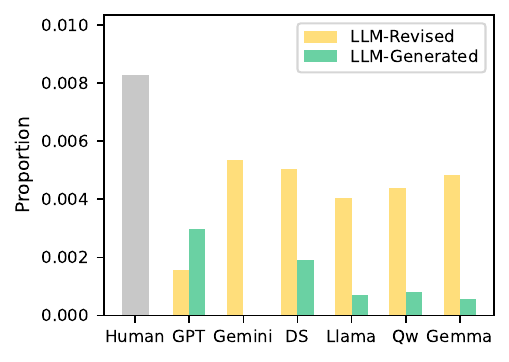}
        \end{subfigure}\hfill
        \begin{subfigure}[t]{0.49\linewidth}
            \centering
            \includegraphics[width=\linewidth]{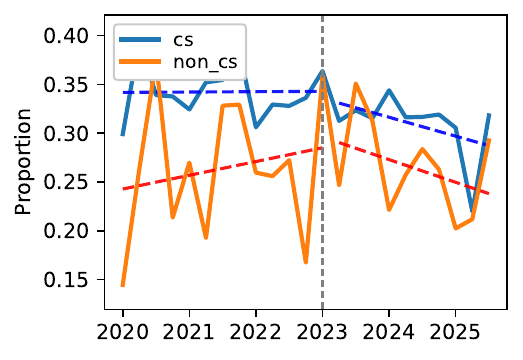}
        \end{subfigure}
        \caption{Other functions.}
    \end{subfigure}

    \begin{subfigure}[t]{0.49\linewidth}
        \centering
        \begin{subfigure}[t]{0.49\linewidth}
            \centering
            \includegraphics[width=\linewidth]{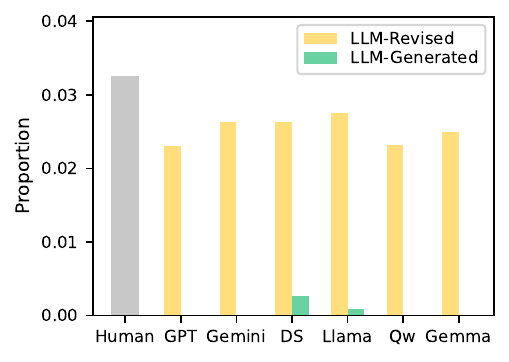}
        \end{subfigure}\hfill
        \begin{subfigure}[t]{0.49\linewidth}
            \centering
            \includegraphics[width=\linewidth]{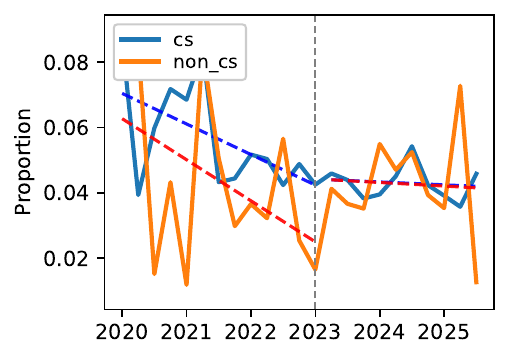}
        \end{subfigure}
        \caption{PascalCase functions.}
    \end{subfigure}\hfill
    \begin{subfigure}[t]{0.49\linewidth}
        \centering
        \begin{subfigure}[t]{0.49\linewidth}
            \centering
            \includegraphics[width=\linewidth]{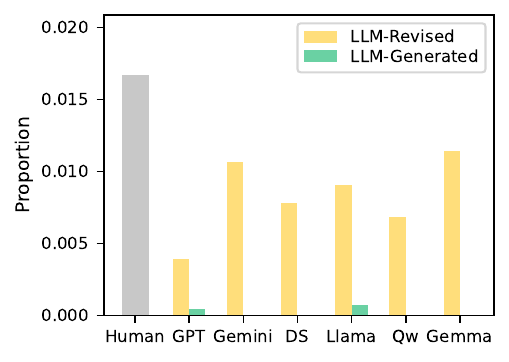}
        \end{subfigure}\hfill
        \begin{subfigure}[t]{0.49\linewidth}
            \centering
            \includegraphics[width=\linewidth]{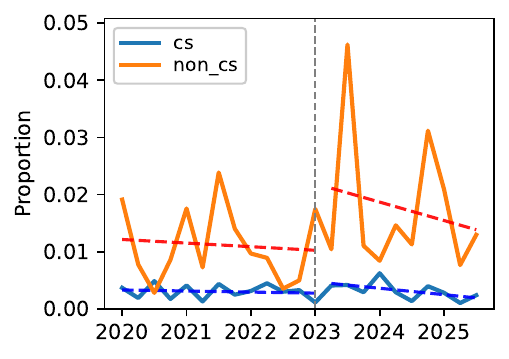}
        \end{subfigure}
        \caption{UPPERCASE functions.}
    \end{subfigure}

    \begin{subfigure}[t]{0.49\linewidth}
        \centering
        \begin{subfigure}[t]{0.49\linewidth}
            \centering
            \includegraphics[width=\linewidth]{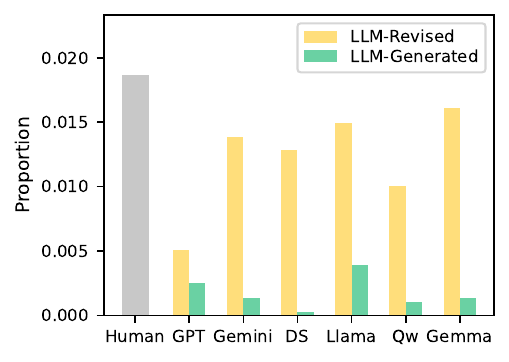}
        \end{subfigure}\hfill
        \begin{subfigure}[t]{0.49\linewidth}
            \centering
            \includegraphics[width=\linewidth]{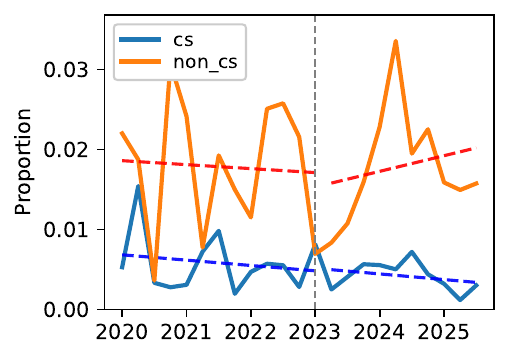}
        \end{subfigure}
        \caption{Single‐letter functions.}
    \end{subfigure}\hfill
    \begin{subfigure}[t]{0.49\linewidth}
        \centering
        \begin{subfigure}[t]{0.49\linewidth}
            \centering
            \includegraphics[width=\linewidth]{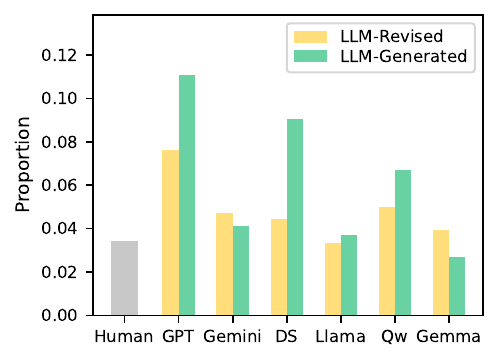}
        \end{subfigure}\hfill
        \begin{subfigure}[t]{0.49\linewidth}
            \centering
            \includegraphics[width=\linewidth]{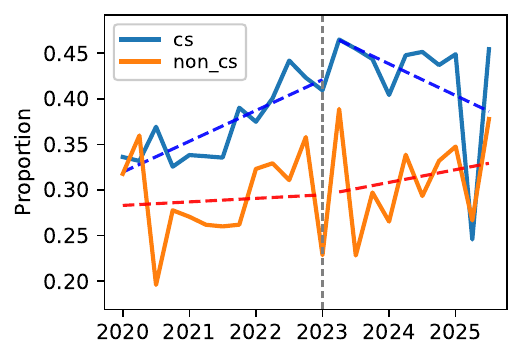}
        \end{subfigure}
        \caption{Snake\_case functions.}
        \label{cpp_func_snake}
    \end{subfigure}
  \begin{subfigure}[t]{0.49\linewidth}
    \centering
    \begin{subfigure}[t]{0.49\linewidth}
      \centering
      \includegraphics[width=\linewidth]{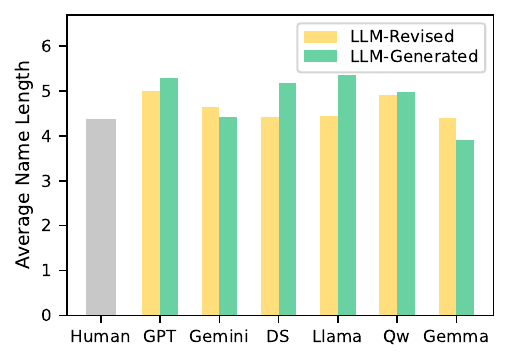}
    \end{subfigure}\hfill
    \begin{subfigure}[t]{0.49\linewidth}
      \centering
      \includegraphics[width=\linewidth]{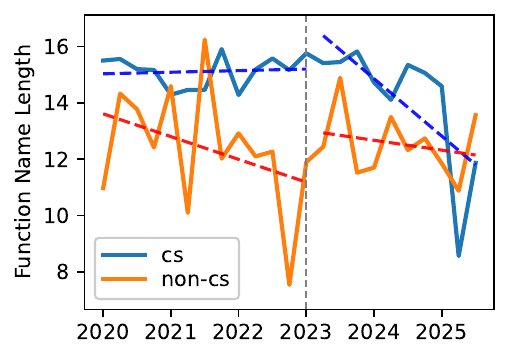}
    \end{subfigure}
    \caption{Average length.}
  \end{subfigure}\hfill
  \begin{subfigure}[t]{0.49\linewidth}
\end{subfigure}

    \caption{Comparison of \textbf{C/C++ function} naming styles in LLM-generated vs.\ human-written code and their temporal trends on GitHub. Analysis reveals that \textbf{LLMs favor snake\_case naming for C/C++ functions}, mirroring a significant increase in the adoption of this convention across GitHub repositories from 2020 to 2025.}
    \label{C/C++ function}
\end{figure*}

\begin{figure*}[!t]
    \centering
    \begin{subfigure}[t]{0.24\linewidth}
        \centering
        \includegraphics[width=\linewidth]{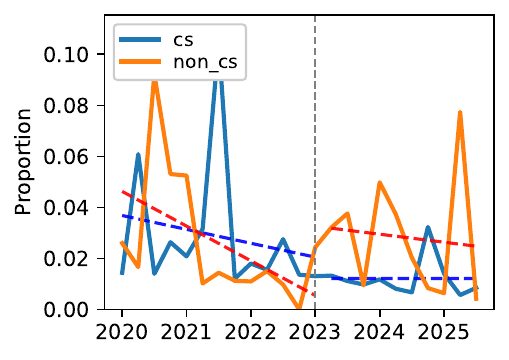}
        \caption{camelCase file names.}
    \end{subfigure}
    \hfill
    \begin{subfigure}[t]{0.24\linewidth}
        \centering
        \includegraphics[width=\linewidth]{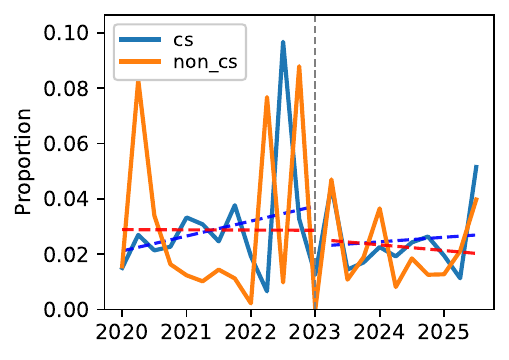}
        \caption{Digit-suffixed file names.}
    \end{subfigure}
    \hfill
    \begin{subfigure}[t]{0.24\linewidth}
        \centering
        \includegraphics[width=\linewidth]{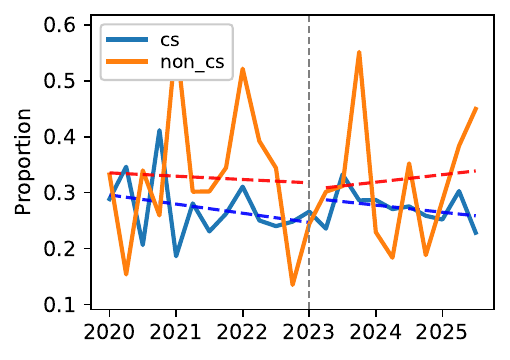}
        \caption{Lowercase file names.}
    \end{subfigure}
    \hfill
    \begin{subfigure}[t]{0.24\linewidth}
        \centering
        \includegraphics[width=\linewidth]{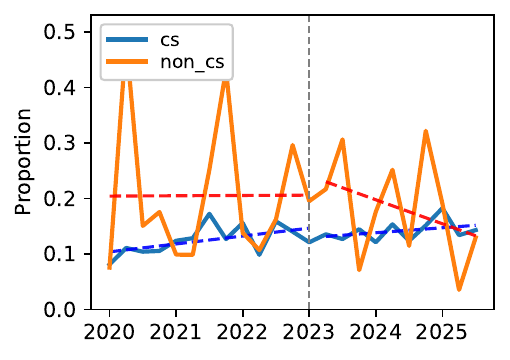}
        \caption{Other file names.}
    \end{subfigure}

    \begin{subfigure}[t]{0.24\linewidth}
        \centering
        \includegraphics[width=\linewidth]{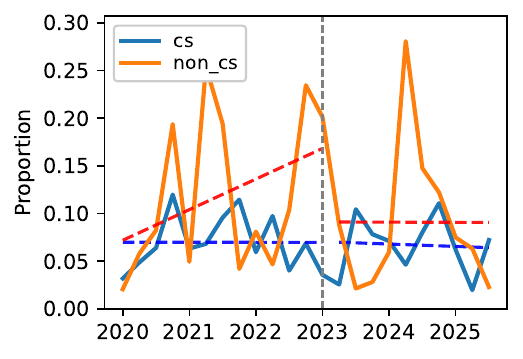}
        \caption{PascalCase file names.}
    \end{subfigure}
    \hfill
    \begin{subfigure}[t]{0.24\linewidth}
        \centering
        \includegraphics[width=\linewidth]{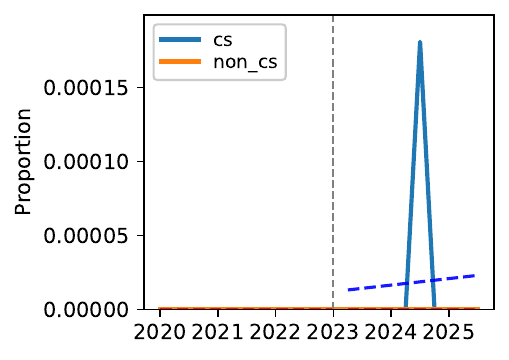}
        \caption{Single letter file names.}
    \end{subfigure}
    \hfill
    \begin{subfigure}[t]{0.24\linewidth}
        \centering
        \includegraphics[width=\linewidth]{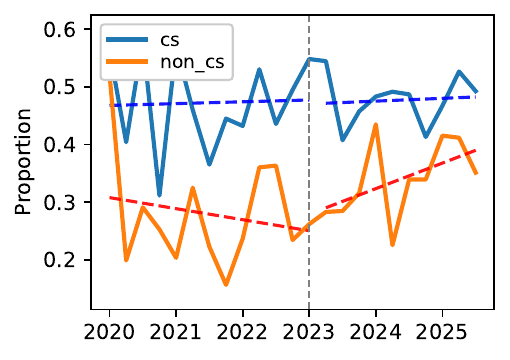}
        \caption{Snake\_case file names.}
    \end{subfigure}
    \hfill
    \begin{subfigure}[t]{0.24\linewidth}
        \centering
        \includegraphics[width=\linewidth]{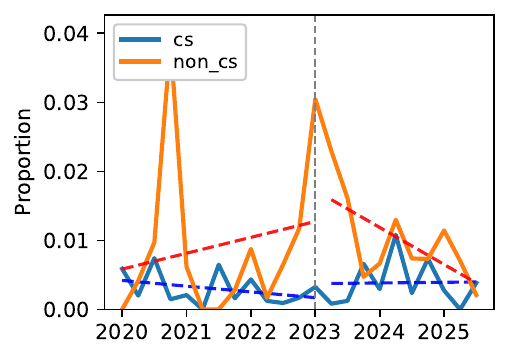}
        \caption{UPPERCASE file names.}
    \end{subfigure}  
    
    \caption{Comparison of \textbf{C/C++ file} naming styles in LLM-generated vs. human-written code and their temporal trends on GitHub. \textbf{All examined naming conventions demonstrate variability without establishing clear trends in the past five years.}}
    \label{C/C++ file}
\end{figure*}

\begin{figure*}[!t]
    \centering
    \begin{subfigure}[t]{0.48\textwidth}
        \centering
        \includegraphics[width=\textwidth]{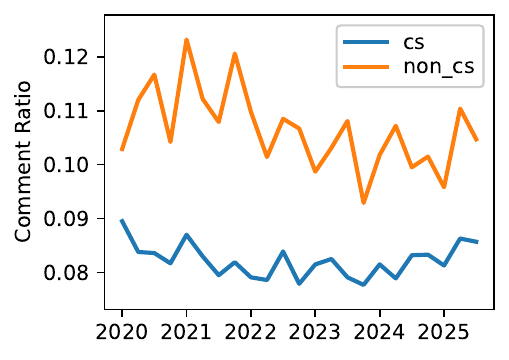}
        \caption{Python comment ratios.}
    \end{subfigure}
    \hfill
    \begin{subfigure}[t]{0.48\textwidth}
        \centering
        \includegraphics[width=\textwidth]{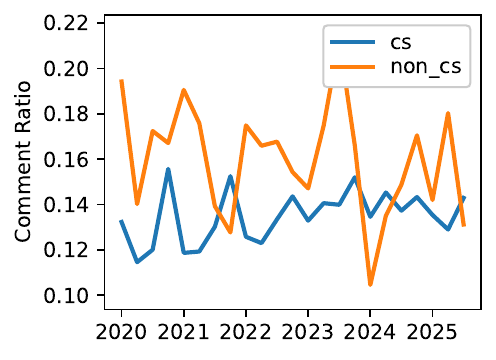}
        \caption{C/C++ comment ratios.}
    \end{subfigure}
    \caption{Unlike the clear patterns observed in naming patterns, \textbf{comment ratios for both languages remain unstable without developing discernible temporal trends.}}
    \label{comments}
\end{figure*}

Figures \ref{Python variable}-\ref{comments} present naming convention preferences of LLMs and corresponding GitHub trends across Python and C/C++ code elements (variables, functions, files) and comment ratios.

\section{Tags Frequencies}
Table~\ref{all_tag} shows the tag frequency of the collected questions. Table~\ref{tag_frequencies} shows the tag frequency results of the model output reasoning.
Table~\ref{tab:match_error_diffi} shows specific results at different levels of difficulty. 
\begin{table*}[!t]
\small
\setlength{\tabcolsep}{2pt} 
\renewcommand{\arraystretch}{1.1}
\centering
\begin{tabular}{l c@{\hskip 10pt}|l c@{\hskip 10pt}|l c@{\hskip 10pt}|l c}
\toprule
Tag & Freq & Tag & Freq & Tag & Freq & Tag & Freq \\
\midrule
2-sat & 7 & binary search & 260 & bitmasks & 81 & brute force & 404 \\
chinese remainder theorem & 3 & combinatorics & 154 & constructive algorithms & 350 & data structures & 395 \\
dfs and similar & 260 & divide and conquer & 57 & dp & 565 & dsu & 91 \\
expression parsing & 26 & fft & 9 & flows & 34 & games & 62 \\
geometry & 157 & graph matchings & 18 & graphs & 312 & greedy & 513 \\
hashing & 55 & implementation & 849 & interactive & 20 & math & 631 \\
matrices & 39 & meet-in-the-middle & 10 & number theory & 184 & probabilities & 78 \\
schedules & 2 & shortest paths & 80 & sortings & 248 & string suffix structures & 25 \\
strings & 221 & ternary search & 15 & trees & 189 & two pointers & 113 \\
\bottomrule
\end{tabular}
\caption{Frequencies of all algorithmic tags.}
\label{all_tag}
\end{table*}

\begin{table*}[!t]
\small
\setlength{\tabcolsep}{3pt} 

\renewcommand{\arraystretch}{1.2}
\centering
\begin{tabular}{l|cccccccccccc}
\toprule
Tag & A\_q\_p & R\_q\_p & A\_q\_c & R\_q\_c & A\_g\_p & R\_g\_p & A\_g\_c & R\_g\_c & A\_d\_p & R\_d\_p & A\_d\_c & R\_d\_c \\
\midrule
2-sat & 0 & 0 & 5 & 3 & 0 & 0 & 2 & 3 & 0 & 0 & 10 & 5 \\
bfs & 35 & 56 & 202 & 151 & 42 & 80 & 228 & 138 & 185 & 133 & 550 & 250 \\
binary search & 132 & 103 & 235 & 363 & 43 & 115 & 73 & 316 & 315 & 152 & 656 & 497 \\
bitmasks & 0 & 1 & 3 & 6 & 1 & 3 & 3 & 6 & 3 & 3 & 3 & 9 \\
brute force & 4 & 0 & 29 & 4 & 5 & 0 & 13 & 0 & 7 & 4 & 25 & 7 \\

chinese remainder & 0 & 0 & 1 & 0 & 0 & 1 & 1 & 0 & 0 & 0 & 3 & 0 \\
combinatorics & 2 & 2 & 9 & 3 & 0 & 0 & 1 & 4 & 1 & 3 & 8 & 5 \\
constructive algorithms & 0 & 0 & 0 & 0 & 0 & 0 & 0 & 0 & 0 & 0 & 0 & 0 \\
data structures & 32 & 10 & 80 & 110 & 3 & 7 & 38 & 90 & 30 & 26 & 69 & 100 \\
dfs & 41 & 41 & 249 & 379 & 63 & 83 & 304 & 431 & 31 & 37 & 183 & 358 \\
divide and conquer & 0 & 0 & 1 & 0 & 0 & 0 & 2 & 2 & 0 & 0 & 12 & 8 \\
dp & 284 & 29 & 1325 & 685 & 325 & 98 & 1497 & 649 & 1145 & 211 & 1977 & 1039 \\
dsu & 1 & 6 & 27 & 15 & 6 & 7 & 9 & 37 & 30 & 8 & 132 & 35 \\
expression parsing & 0 & 0 & 0 & 0 & 0 & 1 & 0 & 0 & 0 & 0 & 0 & 0 \\
fft & 0 & 0 & 0 & 13 & 0 & 0 & 0 & 23 & 0 & 0 & 0 & 13 \\
flows & 0 & 0 & 6 & 2 & 1 & 0 & 9 & 1 & 3 & 1 & 11 & 6 \\
games & 22 & 4 & 23 & 20 & 29 & 16 & 50 & 20 & 16 & 6 & 28 & 18 \\
geometry & 6 & 3 & 19 & 10 & 1 & 1 & 1 & 2 & 2 & 3 & 6 & 5 \\
graph matchings & 0 & 0 & 0 & 0 & 0 & 0 & 0 & 0 & 0 & 0 & 0 & 0 \\
graphs & 7 & 2 & 21 & 10 & 0 & 4 & 11 & 13 & 21 & 3 & 55 & 21 \\
greedy & 65 & 23 & 194 & 41 & 72 & 19 & 227 & 39 & 70 & 26 & 188 & 106 \\
hashing & 0 & 0 & 4 & 16 & 1 & 0 & 0 & 16 & 2 & 1 & 10 & 23 \\
implementation & 180 & 69 & 490 & 212 & 10 & 80 & 52 & 216 & 76 & 59 & 342 & 289 \\
interactive & 2 & 3 & 3 & 3 & 4 & 2 & 6 & 7 & 1 & 2 & 0 & 3 \\
math & 7 & 23 & 1 & 4 & 44 & 16 & 0 & 4 & 89 & 9 & 0 & 0 \\
matrices & 2 & 2 & 7 & 2 & 0 & 1 & 4 & 6 & 6 & 4 & 22 & 16 \\
meet-in-the-middle & 0 & 0 & 1 & 0 & 0 & 0 & 0 & 0 & 0 & 0 & 0 & 0 \\
number theory & 0 & 0 & 3 & 0 & 0 & 0 & 1 & 0 & 1 & 0 & 3 & 3 \\
probabilities & 13 & 25 & 73 & 85 & 9 & 11 & 77 & 77 & 8 & 8 & 105 & 121 \\
schedules & 6 & 2 & 1 & 5 & 0 & 0 & 1 & 1 & 2 & 2 & 1 & 0 \\
shortest paths & 7 & 3 & 57 & 37 & 4 & 9 & 58 & 40 & 8 & 5 & 57 & 63 \\
similar & 18 & 13 & 47 & 31 & 17 & 17 & 45 & 42 & 79 & 17 & 203 & 67 \\
sort & 216 & 63 & 449 & 125 & 271 & 90 & 469 & 131 & 500 & 142 & 645 & 279 \\
string & 696 & 477 & 1154 & 751 & 1021 & 734 & 1498 & 981 & 2366 & 765 & 2296 & 1281 \\
strings & 124 & 121 & 266 & 206 & 201 & 157 & 365 & 205 & 242 & 103 & 412 & 277 \\
suffix & 42 & 28 & 108 & 118 & 45 & 32 & 89 & 98 & 282 & 47 & 272 & 164 \\
ternary search & 0 & 0 & 0 & 12 & 0 & 0 & 3 & 11 & 1 & 0 & 3 & 16 \\
trees & 19 & 11 & 107 & 92 & 14 & 14 & 101 & 101 & 36 & 16 & 209 & 140 \\
two pointers & 25 & 15 & 52 & 14 & 5 & 3 & 10 & 5 & 14 & 7 & 24 & 12 \\

\bottomrule
\end{tabular}

\caption{Frequencies of each output tag across different generation types. The format \texttt{T\_M\_L} denotes \textbf{Type} (A: ANS, R: REF), \textbf{Model} (d: \textit{DeepSeek-R1-Distill-Qwen-32B}, q: \textit{Qwen3-32B}, g: \textit{Gemma-3-27B}), and \textbf{Language} (p: Python, c: C/C++).}

\label{tag_frequencies}
\end{table*}

\begin{table*}[!t]
\setlength{\tabcolsep}{2pt} 
\small
\renewcommand{\arraystretch}{1.2}
\centering
\begin{tabular}{l|cccccccccccc}
\toprule
Tag & A\_gp\_p & R\_gp\_p & A\_gp\_c & R\_gp\_c & A\_ge\_p & R\_ge\_p & A\_ge\_c & R\_ge\_c & A\_l\_p & R\_l\_p & A\_l\_c & R\_l\_c \\
\midrule
2-sat & 0 & 0 & 15 & 4 & 4 & 1 & 7 & 7 & 0 & 0 & 24 & 5 \\
bfs & 94 & 95 & 509 & 182 & 44 & 49 & 196 & 92 & 48 & 49 & 176 & 120 \\
binary search & 142 & 137 & 551 & 361 & 73 & 97 & 184 & 243 & 72 & 95 & 160 & 288 \\
bitmasks & 3 & 4 & 21 & 11 & 1 & 1 & 6 & 7 & 0 & 0 & 2 & 3 \\
brute force & 43 & 15 & 108 & 23 & 6 & 2 & 29 & 5 & 3 & 1 & 10 & 1 \\
chinese remainder & 0 & 0 & 3 & 3 & 0 & 0 & 3 & 1 & 1 & 1 & 2 & 1 \\
combinatorics & 0 & 3 & 15 & 24 & 0 & 1 & 3 & 6 & 1 & 1 & 4 & 7 \\
constructive algorithms & 0 & 0 & 1 & 0 & 0 & 0 & 0 & 0 & 0 & 0 & 0 & 0 \\
data structures & 15 & 13 & 96 & 101 & 11 & 4 & 55 & 58 & 16 & 10 & 74 & 73 \\
dfs & 94 & 61 & 408 & 369 & 71 & 58 & 332 & 355 & 50 & 49 & 277 & 411 \\
divide and conquer & 0 & 0 & 26 & 6 & 0 & 2 & 2 & 6 & 0 & 0 & 0 & 3 \\
dp & 779 & 104 & 4230 & 1130 & 399 & 82 & 2172 & 764 & 87 & 52 & 752 & 802 \\
dsu & 2 & 7 & 103 & 10 & 3 & 7 & 19 & 17 & 0 & 3 & 0 & 8 \\
expression parsing & 0 & 0 & 0 & 0 & 0 & 0 & 0 & 0 & 1 & 0 & 0 & 0 \\
fft & 0 & 0 & 15 & 19 & 0 & 0 & 0 & 24 & 0 & 0 & 0 & 10 \\
flows & 3 & 2 & 35 & 8 & 1 & 1 & 3 & 0 & 0 & 1 & 0 & 1 \\
games & 68 & 12 & 67 & 21 & 17 & 15 & 41 & 16 & 39 & 31 & 66 & 31 \\
geometry & 3 & 2 & 16 & 12 & 1 & 1 & 1 & 1 & 13 & 1 & 31 & 5 \\
graph matchings & 0 & 0 & 0 & 0 & 0 & 0 & 0 & 0 & 0 & 0 & 0 & 0 \\
graphs & 10 & 7 & 45 & 19 & 1 & 5 & 10 & 9 & 3 & 2 & 37 & 15 \\
greedy & 79 & 27 & 252 & 52 & 39 & 20 & 127 & 56 & 9 & 5 & 49 & 26 \\
hashing & 4 & 1 & 31 & 27 & 0 & 0 & 0 & 12 & 3 & 0 & 1 & 13 \\
implementation & 202 & 42 & 1006 & 157 & 12 & 41 & 58 & 91 & 42 & 10 & 228 & 48 \\
interactive & 5 & 6 & 5 & 10 & 3 & 2 & 2 & 3 & 0 & 2 & 0 & 3 \\
math & 16 & 72 & 15 & 19 & 9 & 17 & 9 & 2 & 3 & 5 & 2 & 2 \\
matrices & 8 & 5 & 54 & 13 & 1 & 0 & 6 & 5 & 2 & 1 & 3 & 5 \\
meet-in-the-middle & 0 & 0 & 9 & 3 & 0 & 0 & 0 & 1 & 0 & 0 & 0 & 1 \\
number theory & 1 & 0 & 4 & 7 & 0 & 1 & 0 & 0 & 0 & 1 & 2 & 1 \\
probabilities & 11 & 10 & 114 & 58 & 2 & 5 & 34 & 48 & 14 & 8 & 109 & 104 \\
schedules & 1 & 2 & 2 & 4 & 1 & 0 & 1 & 2 & 1 & 1 & 1 & 0 \\
shortest paths & 12 & 10 & 102 & 39 & 11 & 4 & 46 & 26 & 11 & 6 & 43 & 19 \\
similar & 99 & 33 & 392 & 72 & 17 & 20 & 52 & 40 & 39 & 20 & 163 & 41 \\
sort & 225 & 112 & 816 & 205 & 214 & 59 & 455 & 122 & 183 & 79 & 414 & 115 \\
string & 880 & 599 & 1603 & 924 & 819 & 511 & 1280 & 790 & 711 & 739 & 1158 & 1210 \\
strings & 182 & 137 & 525 & 236 & 175 & 132 & 316 & 154 & 142 & 130 & 313 & 252 \\
suffix & 293 & 104 & 712 & 206 & 43 & 42 & 118 & 86 & 46 & 23 & 140 & 97 \\
ternary search & 2 & 0 & 9 & 19 & 0 & 0 & 0 & 13 & 5 & 0 & 0 & 14 \\
trees & 28 & 19 & 305 & 125 & 12 & 8 & 81 & 79 & 15 & 15 & 140 & 133 \\
two pointers & 21 & 13 & 112 & 17 & 5 & 7 & 15 & 4 & 9 & 15 & 16 & 16 \\

\bottomrule
\end{tabular}

\caption{Frequencies of each output tag across different generation types. The format \texttt{T\_M\_L} denotes \textbf{Type} (A: ANS, R: REF), \textbf{Model} (gp: \textit{GPT-4.1}, ge: \textit{Gemini-2.0-flash},  l: \textit{Llama-4-Maverick}), and \textbf{Language} (p: Python, c: C/C++).}

\label{tag_frequencies_continue}
\end{table*}

\begin{table*}[!t]
\small
\setlength{\tabcolsep}{2pt}
\renewcommand{\arraystretch}{1.5}
\centering
\begin{tabular}{l|cc|cc|cc|cc}
\toprule
\textbf{Model-Language-Type} 
& \multicolumn{2}{c|}{\textbf{800–1199}} 
& \multicolumn{2}{c|}{\textbf{1200–1599}} 
& \multicolumn{2}{c|}{\textbf{1600–1999}} 
& \multicolumn{2}{c}{\textbf{2000+}} \\
\cmidrule(lr){2-3} \cmidrule(lr){4-5} \cmidrule(lr){6-7} \cmidrule(lr){8-9}
& Match Rate & Error Rate & Match Rate & Error Rate & Match Rate & Error Rate & Match Rate & Error Rate \\
\midrule
Qwen\_Python\_ans & 9.24\% & 17.39\% & 18.44\% & 23.12\% & 24.84\% & 34.28\% & 27.73\% & 33.61\% \\
Qwen\_Python\_ref & 9.24\% & 11.68\% & 11.43\% & 15.58\% & 16.67\% & 20.13\% & 19.33\% & 25.21\% \\
Qwen\_cpp\_ans & 13.73\% & 19.95\% & 21.26\% & 22.43\% & 30.30\% & 35.90\% & 33.37\% & 44.85\% \\
Qwen\_cpp\_ref & 9.33\% & 11.66\% & 12.85\% & 18.69\% & 27.15\% & 21.54\% & 38.29\% & 33.26\% \\
Gemma\_Python\_ans & 5.43\% & 10.60\% & 9.87\% & 18.44\% & 15.41\% & 20.13\% & 15.97\% & 26.89\% \\
Gemma\_Python\_ref & 6.25\% & 9.51\% & 11.17\% & 13.25\% & 20.44\% & 21.07\% & 20.17\% & 30.25\% \\
Gemma\_cpp\_ans & 5.70\% & 12.95\% & 11.68\% & 17.06\% & 18.04\% & 18.21\% & 21.55\% & 25.29\% \\
Gemma\_cpp\_ref & 8.03\% & 8.55\% & 12.62\% & 15.65\% & 27.15\% & 18.21\% & 33.37\% & 31.62\% \\
Deepseek\_Python\_ans & 5.43\% & 13.59\% & 15.32\% & 20.52\% & 23.90\% & 23.27\% & 17.65\% & 30.25\% \\
Deepseek\_Python\_ref & 5.43\% & 10.87\% & 12.73\% & 11.43\% & 18.55\% & 18.55\% & 22.69\% & 28.57\% \\
Deepseek\_cpp\_ans & 11.92\% & 16.06\% & 20.56\% & 24.30\% & 33.27\% & 28.20\% & 35.48\% & 41.92\% \\
Deepseek\_cpp\_ref & 10.88\% & 13.47\% & 20.79\% & 19.63\% & 32.92\% & 25.74\% & 43.79\% & 34.31\% \\
GPT\_Python\_ans & 6.25\% & 14.13\% & 22.34\% & 27.79\% & 35.22\% & 36.48\% & 40.34\% & 61.34\% \\
GPT\_Python\_ref & 7.07\% & 14.13\% & 13.51\% & 23.12\% & 27.04\% & 28.62\% & 26.89\% & 36.97\% \\
GPT\_cpp\_ans & 13.21\% & 15.80\% & 27.57\% & 29.44\% & 45.18\% & 48.16\% & 50.00\% & 76.35\% \\
GPT\_cpp\_ref & 8.03\% & 14.77\% & 14.95\% & 21.50\% & 32.05\% & 23.99\% & 45.55\% & 38.64\% \\
Gemini\_Python\_ans & 4.62\% & 10.33\% & 10.13\% & 14.81\% & 21.07\% & 19.50\% & 22.69\% & 24.37\% \\
Gemini\_Python\_ref & 5.98\% & 8.70\% & 10.65\% & 13.51\% & 19.50\% & 11.95\% & 17.65\% & 21.01\% \\
Gemini\_cpp\_ans & 5.44\% & 10.36\% & 10.05\% & 16.12\% & 23.99\% & 20.32\% & 33.49\% & 30.91\% \\
Gemini\_cpp\_ref & 6.48\% & 8.29\% & 11.92\% & 10.75\% & 26.44\% & 13.31\% & 33.02\% & 24.24\% \\
Llama4\_Python\_ans & 1.90\% & 11.96\% & 10.91\% & 14.29\% & 21.07\% & 16.67\% & 23.53\% & 27.73\% \\
Llama4\_Python\_ref & 4.08\% & 7.88\% & 9.35\% & 13.25\% & 14.47\% & 13.21\% & 17.65\% & 21.01\% \\
Llama4\_cpp\_ans & 3.89\% & 12.69\% & 13.55\% & 14.72\% & 26.62\% & 17.69\% & 28.92\% & 36.89\% \\
Llama4\_cpp\_ref & 6.22\% & 10.36\% & 12.15\% & 14.25\% & 27.15\% & 14.89\% & 34.89\% & 27.99\% \\
\bottomrule
\end{tabular}
\caption{Match rate and error rate for questions across different difficulty levels.}
\label{tab:match_error_diffi}
\end{table*}

\section{Metrics Result}
Tables~\ref{tab:halstead_stats} and~\ref{tab:maintainability_stats} show the results containing reference-guided generation.
Tables~\ref{tab:halstead_status_subset} and~\ref{tab:maintainability_status_subset} show the results of large-scale model evaluation.

\begin{table*}[!t]
\small
\setlength{\tabcolsep}{3pt} 

\renewcommand{\arraystretch}{1.5}
\centering

\begin{adjustbox}{center}
\begin{tabular}{l|cccccccccccc}
\toprule
\textbf{Model}             & \textbf{n1}    & \textbf{n2}     & \textbf{N1}     & \textbf{N2}     & \textbf{Vocab} & \textbf{Length} & \textbf{Cal\_Len} & \textbf{Volume}   & \textbf{Difficul} & \textbf{Effort}    & \textbf{Time\_Sec} & \textbf{Bugs}   \\
\midrule
Qwen\_ANS\_Python & 5 & 14.64 & 11.9 & 23.09 & 19.64 & 35 & 73.89 & 162.62 & 3.93 & 909.43 & 50.52 & 0.05 \\
Qwen\_REF\_Python & 4.97 & 14.27 & 11.47 & 22.25 & 19.24 & 33.72 & 72.27 & 157.44 & 3.88 & 921.69 & 51.21 & 0.05 \\
Gemma\_ANS\_Python & 4.8 & 14.98 & 12.21 & 24.13 & 19.78 & 36.33 & 76.08 & 172.55 & 3.81 & 982.84 & 54.6 & 0.06 \\
Gemma\_REF\_Python & 4.95 & 14.34 & 11.56 & 22.42 & 19.29 & 33.99 & 72.63 & 158.69 & 3.88 & 923.77 & 51.32 & 0.05 \\
Deepseek\_ANS\_Python & 5.57 & 16.15 & 14.04 & 27.05 & 21.72 & 41.09 & 85.91 & 201.03 & 4.63 & 1472.81 & 81.82 & 0.07 \\
Deepseek\_REF\_Python & 5.59 & 16.23 & 13.67 & 26.37 & 21.82 & 40.04 & 86.31 & 195.18 & 4.54 & 1345.86 & 74.77 & 0.07 \\
GPT\_ANS\_Python & 5.44 & 15.64 & 12.7 & 24.49 & 21.08 & 37.19 & 82.35 & 179.33 & 4.25 & 1168.85 & 64.94 & 0.06 \\
GPT\_REF\_Python & 5.13 & 14.64 & 12.06 & 23.29 & 19.77 & 35.36 & 75.55 & 168.17 & 4.07 & 1092.46 & 60.69 & 0.06 \\
Gemini\_ANS\_Python & 4.96 & 14.71 & 12.62 & 24.49 & 19.66 & 37.11 & 74.93 & 176.87 & 4.04 & 1202.63 & 66.81 & 0.06 \\
Gemini\_REF\_Python & 5.1 & 15.34 & 12.54 & 24.36 & 20.44 & 36.9 & 79.31 & 176.79 & 4.04 & 1090.5 & 60.58 & 0.06 \\
Llama4\_ANS\_Python & 5.43 & 16.36 & 13.08 & 25.5 & 21.79 & 38.58 & 84.89 & 184.97 & 4.18 & 1113.48 & 61.86 & 0.06 \\
Llama4\_REF\_Python & 5.25 & 16.2 & 13.38 & 25.96 & 21.45 & 39.34 & 85.19 & 192.17 & 4.18 & 1259.67 & 69.98 & 0.06 \\
AC\_Python & 5.31 & 17.02 & 14.23 & 27.69 & 22.33 & 41.92 & 90.1 & 206.39 & 4.32 & 1353.9 & 75.22 & 0.07 \\
\midrule
Qwen\_ANS\_C++ & 8.99 & 54.75 & 78.42 & 258.06 & 63.73 & 336.48 & 349.75 & 2070.25 & 20.73 & 58310.92 & 3239.5 & 0.69 \\
Qwen\_REF\_C++ & 9.46 & 56.44 & 87.92 & 332.05 & 65.9 & 419.96 & 365.58 & 2632.6 & 26.59 & 105535.64 & 5863.09 & 0.88 \\
Gemma\_ANS\_C++ & 7.88 & 45.9 & 92 & 251.6 & 53.78 & 343.6 & 279.63 & 2018.25 & 21.32 & 56967.93 & 3164.88 & 0.67 \\
Gemma\_REF\_C++ & 9.33 & 63.15 & 102.18 & 372.22 & 72.48 & 474.4 & 418.74 & 3064.43 & 26.34 & 122666.91 & 6814.83 & 1.02 \\
Deepseek\_ANS\_C++ & 8.75 & 53.2 & 87.53 & 292.85 & 61.95 & 380.38 & 337.54 & 2333.53 & 23.4 & 74950.39 & 4163.91 & 0.78 \\
Deepseek\_REF\_C++ & 9.48 & 57.28 & 92.37 & 357.19 & 66.76 & 449.56 & 372.17 & 2824.23 & 28.23 & 117793.48 & 6544.08 & 0.94 \\
GPT\_ANS\_C++ & 9.66 & 89.02 & 128.68 & 433.18 & 98.68 & 561.86 & 631.73 & 3953.33 & 21.76 & 123942.64 & 6885.7 & 1.32 \\
GPT\_REF\_C++ & 9.79 & 72.81 & 103.59 & 386.58 & 82.61 & 490.17 & 495.58 & 3253.06 & 24.9 & 115793.09 & 6432.95 & 1.08 \\
Gemini\_ANS\_C++ & 8.04 & 45.78 & 82.66 & 259.71 & 53.82 & 342.37 & 280.16 & 2022.3 & 22.34 & 63240.49 & 3513.36 & 0.67 \\
Gemini\_REF\_C++ & 9.37 & 64.07 & 103.19 & 382.94 & 73.43 & 486.13 & 426.49 & 3147.35 & 27 & 124454.36 & 6914.13 & 1.05 \\
Llama4\_ANS\_C++ & 8.73 & 49.13 & 83.51 & 265.57 & 57.87 & 349.08 & 307.25 & 2090.2 & 23.19 & 63666.78 & 3537.04 & 0.7 \\
Llama4\_REF\_C++ & 9.47 & 56.54 & 90.92 & 355.35 & 66.01 & 446.27 & 366.4 & 2795.06 & 28.41 & 117857.74 & 6547.65 & 0.93 \\
AC\_C++ & 10.11 & 72.93 & 123.9 & 458.05 & 83.03 & 581.94 & 497.13 & 3857.01 & 30.67 & 170622.85 & 9479.05 & 1.29 \\
\bottomrule
\end{tabular}

\end{adjustbox}
\caption{
Halstead results. Each label follows the format \textit{model\_type\_language}, where \textit{type} refers to the experimental setting and \textit{language} indicates the programming language. Metric abbreviations: \texttt{cal\_len} (calculated program length), \texttt{difficul}(difficulty), \texttt{time\_sec}(time to implement), \texttt{bugs}(estimated bugs).
}

\label{tab:halstead_stats}
\end{table*}

\begin{table*}[!t]
\small
\renewcommand{\arraystretch}{1.5}
\centering
\begin{tabular}{lcccccc}
\toprule
\textbf{Model}   & \textbf{Problem}   & \textbf{Submission\_id} & \textbf{Type}      & \textbf{State} & \textbf{Time} & \textbf{Memory} \\
\midrule
Qwen & 584B & 19821& ANS	& WA Test1 & - & - \\
& 584B & 19821& REF	& AC & 61ms & 52KB \\
& 586A & 35393& ANS	& WA Test2 & - & - \\
& 586A & 35393& REF	& WA Test7 & - & - \\
& 584A & 19768& ANS	& WA Test7 & - & - \\
& 584A & 19768& REF	& WA Test10 & - & - \\
\midrule
Gemma & 584B & 19821& ANS	& WA Test3 & - & - \\
& 584B & 19821& REF	& WA Test3 & - & - \\
& 586A & 35393& ANS	& WA Test1 & - & - \\
& 586A & 35393& REF	& WA Test2 & - & - \\
& 584A & 19768& ANS	& CE & - & - \\
& 584A & 19768& REF	& AC & 62ms & 40KB \\
\midrule
Deepseek & 584B & 19821& ANS	& AC & 77ms & 48KB \\
& 584B & 19821& REF	& AC & 62ms & 48KB \\
& 586A & 35393& ANS	& WA Test2 & - & - \\
& 586A & 35393& REF	& AC & 62ms & 44KB \\
& 584A & 19768& ANS	& AC & 77ms & 44KB \\
& 584A & 19768& REF	& WA Test10 & - & - \\
\midrule
GPT & 584B & 19821& ANS	& AC & 62ms & 56KB \\
& 584B & 19821& REF	& AC & 62ms & 40KB \\
& 586A & 35393& ANS	& WA Test2 & - & - \\
& 586A & 35393& REF	& WA Test2 & - & - \\
& 584A & 19768& ANS	& AC & 62ms & 60KB \\
& 584A & 19768& REF	& AC & 61ms & 44KB \\
\midrule
Gemini & 584B & 19821& ANS	& AC & 61ms & 52KB \\
& 584B & 19821& REF	& AC & 62ms & 48KB \\
& 586A & 35393& ANS	& WA Test2 & - & - \\
& 586A & 35393& REF	& WA Test2 & - & - \\
& 584A & 19768& ANS	& AC & 62ms & 56KB \\
& 584A & 19768& REF	& AC & 62ms & 52KB \\
\midrule
Llama4 & 584B & 19821& ANS	& AC & 61ms & 56KB \\
& 584B & 19821& REF	& AC & 46ms & 52KB \\
& 586A & 35393& ANS	& WA Test2 & - & - \\
& 586A & 35393& REF	& AC & 62ms & 64KB \\
& 584A & 19768& ANS	& WA Test10 & - & - \\
& 584A & 19768& REF	& WA Test12 & - & - \\
\midrule
Human & 584B & 19821& AC	& AC & 78ms & 44KB \\
& 586A & 35393& AC	& AC & 61ms & 52KB \\
& 584A & 19768& REF	& AC & 62ms & 832KB \\
\bottomrule
\end{tabular}
\caption{
Case Study on testing LLM-generated code: We selected problems 584B, 586A, and 584A (corresponding to submission IDs 19821, 39853, and 19768, respectively) and evaluated their correctness (compiled in C++17 (GCC 7-32)). \textit{AC}C represents Accept, \textit{WA TestX} represents a Wrong Answer in the Xth test set, and \textit{CE} represents a Compilation Error. For correct code, \textit{Time} and \textit{Memory} represent the time and memory consumption.
}

\label{tab:accuracy_case}
\end{table*}

\begin{table*}[!t]
\small
\renewcommand{\arraystretch}{1.5}
\centering
\begin{tabular}{l|ccccccc}
\toprule
\textbf{Model}             & \textbf{Volume}   & \textbf{Cyclomatic} & \textbf{SLOC}   & \textbf{LLOC}   & \textbf{Comment\_Rate} & \textbf{mi\_std} & \textbf{mi\_custom} \\
\midrule
Qwen\_ANS\_Python & 162.62 & 6.27 & 21.63 & 22.09 & 22.12 & 77.29 & 77.34 \\
Qwen\_REF\_Python & 157.44 & 2.64 & 17.42 & 17.7 & 5.75 & 66.15 & 66.65 \\
Gemma\_ANS\_Python & 172.55 & 7.94 & 22.39 & 22.47 & 0.95 & 58.99 & 58.95 \\
Gemma\_REF\_Python & 158.69 & 2.91 & 18.56 & 18.98 & 2.13 & 62.63 & 63.1 \\
Deepseek\_ANS\_Python & 201.03 & 2.49 & 24.82 & 24.95 & 4.6 & 62 & 62.79 \\
Deepseek\_REF\_Python & 195.18 & 3.08 & 22.16 & 22.4 & 5.79 & 63.84 & 64.48 \\
GPT\_ANS\_Python & 179.33 & 3.03 & 22.15 & 22.19 & 14.74 & 72.22 & 72.81 \\
GPT\_REF\_Python & 168.17 & 2 & 18.66 & 18.79 & 8.17 & 69.08 & 69.7 \\
Gemini\_ANS\_Python & 176.87 & 6.25 & 22.59 & 22.81 & 1.04 & 59.55 & 59.67 \\
Gemini\_REF\_Python & 176.79 & 2.66 & 18.98 & 19.47 & 2.16 & 62.32 & 62.8 \\
Llama4\_ANS\_Python & 184.97 & 7.56 & 22.09 & 22.42 & 3.85 & 60.43 & 60.47 \\
Llama4\_REF\_Python & 192.17 & 2.92 & 18.98 & 19.46 & 0.83 & 60.48 & 61.06 \\
AC\_Python & 206.39 & 2.53 & 20.04 & 20.82 & 3.66 & 62.58 & 62.96    \\\midrule
Qwen\_ANS\_C++ & 2070.25 & 9.97 & 40.13 & 35.12 & 0.04 & 49.66 & 42.35 \\
Qwen\_REF\_C++ & 2632.6 & 11.89 & 46.3 & 38.5 & 0.01 & 43 & 40.82 \\
Gemma\_ANS\_C++ & 2018.25 & 13.49 & 44.01 & 39.69 & 0.01 & 42.41 & 41.03 \\
Gemma\_REF\_C++ & 3064.43 & 12.77 & 52.82 & 40.68 & 0.02 & 43.06 & 39.27 \\
Deepseek\_ANS\_C++ & 2333.53 & 11.52 & 48.26 & 42.53 & 0.01 & 42.79 & 40.33 \\
Deepseek\_REF\_C++ & 2824.23 & 12.68 & 49.72 & 41.09 & 0.01 & 41.21 & 39.9 \\
GPT\_ANS\_C++ & 3953.33 & 13.03 & 59.76 & 46.17 & 0.12 & 53.25 & 38.18 \\
GPT\_REF\_C++ & 3253.06 & 12.98 & 53.04 & 43.47 & 0.06 & 48.47 & 39.12 \\
Gemini\_ANS\_C++ & 2022.3 & 11.21 & 44.06 & 39.85 & 0.01 & 43.12 & 41.7 \\
Gemini\_REF\_C++ & 3147.35 & 12.54 & 53.58 & 41.2 & 0.03 & 43.13 & 39.14 \\
Llama4\_ANS\_C++ & 2090.2 & 10.39 & 40.96 & 35.86 & 0.01 & 44.02 & 41.99 \\
Llama4\_REF\_C++ & 2795.06 & 12.39 & 47.11 & 38.48 & 0 & 41.01 & 40.48 \\
AC\_C++ & 3857.01 & 13.93 & 61.6 & 44.03 & 0.05 & 43.27 & 37.07 \\
\bottomrule
\end{tabular}
\caption{
Maintainability results. Each label follows the format \textit{model\_type\_language}, where \textit{type} refers to the experimental setting and \textit{language} indicates the programming language. Metric abbreviations: \texttt{mi\_std} (standard maintainability index), \texttt{mi\_custom} (custom maintainability index).
}

\label{tab:maintainability_stats}
\end{table*}

\begin{table*}[!t]
\small
\setlength{\tabcolsep}{3.5pt}
\renewcommand{\arraystretch}{1.05}
\centering

\begin{adjustbox}{center}
\begin{tabular}{l l l|ccccccccccccc}
\toprule
\textbf{Model} & \textbf{Language} & \textbf{Stat.} 
 & \textbf{n1} & \textbf{n2} & \textbf{N\textsubscript{1}} & \textbf{N\textsubscript{2}} & \textbf{Vocab} & \textbf{Length} & \textbf{Cal\_Len} & \textbf{Volume} & \textbf{Difficul} & \textbf{Effort} & \textbf{Time\_Sec} & \textbf{Bugs} \\
\midrule
\multirow{4}{*}{\textbf{Claude}} 
 & \multirow{2}{*}{Python} & mean 
    & 6.22 & 20.36 & 18.01 & 35.13 & 26.58 & 53.14 & 114.86 & 277.88 & 5.30 & 2303.85 & 127.99 & 0.09 \\
 &                         & std  
    & 1.31 &  8.14 &  8.26 & 16.27 &  9.07 & 24.51 &  58.33 & 164.09 & 1.49 & 2061.31 & 114.52 & 0.05 \\
 \cmidrule(lr){2-15}
 & \multirow{2}{*}{C/C++}  & mean 
    & 8.54 & 65.90 & 96.68 & 312.96 & 74.44 & 409.64 & 430.84 & 2611.66 & 19.76 & 67162.33 & 3731.24 & 0.87 \\
 &                         & std  
    & 0.91 & 11.60 & 23.72 &  75.44 & 11.94 &  96.36 &  89.76 &  698.95 &  4.31 & 34582.01 & 1921.22 & 0.23 \\
\midrule
\multirow{4}{*}{\textbf{DSV3}} 
 & \multirow{2}{*}{Python} & mean 
    & 6.10 & 18.37 & 17.23 & 33.28 & 24.46 & 50.52 & 100.73 & 254.75 & 5.49 & 2102.24 & 116.79 & 0.08 \\
 &                         & std  
    & 0.77 &  4.14 &  4.58 &  8.93 &  4.65 & 13.50 &  28.68 &  85.88 & 0.99 & 1111.11 &  61.73 & 0.03 \\
 \cmidrule(lr){2-15}
 & \multirow{2}{*}{C/C++}  & mean 
    & 7.65 & 48.08 & 71.45 & 249.61 & 55.73 & 321.06 & 294.59 & 1908.68 & 19.52 & 51055.60 & 2836.42 & 0.64 \\
 &                         & std  
    & 0.62 &  3.52 & 10.80 &  36.49 &  3.81 &  46.29 &  26.88 &  305.03 & 2.99 & 18789.07 & 1043.84 & 0.10 \\
\midrule
\multirow{4}{*}{\textbf{DSR1}} 
 & \multirow{2}{*}{Python} & mean 
    & 6.84 & 22.29 & 20.54 & 39.59 & 29.12 & 60.13 & 128.84 & 319.45 & 6.05 & 2902.90 & 161.27 & 0.11 \\
 &                         & std  
    & 1.23 &  7.26 &  7.64 & 14.87 &  8.13 & 22.49 &  51.49 & 147.19 & 1.47 & 1927.68 & 107.09 & 0.05 \\
 \cmidrule(lr){2-15}
 & \multirow{2}{*}{C/C++}  & mean 
    & 8.42 & 57.94 & 84.00 & 337.67 & 66.36 & 421.67 & 377.04 & 2720.12 & 22.66 & 98184.66 & 5454.70 & 0.91 \\
 &                         & std  
    & 0.88 & 18.49 & 31.58 & 184.28 & 18.88 & 213.40 & 158.33 & 1753.39 & 5.24 & 102703.88 & 5705.77 & 0.58 \\
\midrule
\multirow{4}{*}{\textbf{Gemma}} 
 & \multirow{2}{*}{Python} & mean 
    & 5.52 & 21.20 & 17.90 & 36.14 & 26.72 & 54.04 & 115.75 & 278.83 & 4.60 & 1792.06 &  99.56 & 0.09 \\
 &                         & std  
    & 0.92 &  6.32 &  5.92 & 12.28 &  6.86 & 18.17 &  42.99 & 117.70 & 0.98 & 1076.68 &  59.82 & 0.04 \\
 \cmidrule(lr){2-15}
 & \multirow{2}{*}{C/C++}  & mean 
    & 7.11 & 47.51 & 112.19 & 293.85 & 54.62 & 406.04 & 287.21 & 2379.33 & 21.96 & 67879.88 & 3771.10 & 0.79 \\
 &                         & std  
    & 0.65 &  3.51 &  24.80 &  53.41 &  3.81 &  76.19 &  26.26 &  474.05 & 4.14 & 26880.38 & 1493.35 & 0.16 \\
\midrule
\multirow{4}{*}{\textbf{GPT}} 
 & \multirow{2}{*}{Python} & mean 
    & 5.01 & 13.60 & 11.32 & 22.04 & 18.62 & 33.36 &  67.50 & 151.65 & 4.07 &  912.18 &  50.68 & 0.05 \\
 &                         & std  
    & 1.02 &  3.80 &  3.58 &  7.07 &  4.53 & 10.64 &  23.91 &  59.85 & 1.07 &  528.82 &  29.38 & 0.02 \\
 \cmidrule(lr){2-15}
 & \multirow{2}{*}{C/C++}  & mean 
    & 7.59 & 48.34 & 63.12 & 203.35 & 55.93 & 266.48 & 295.94 & 1571.22 & 15.88 & 30078.52 & 1671.03 & 0.52 \\
 &                         & std  
    & 0.93 &  7.75 & 11.30 &  32.64 &  8.08 &  41.75 &  57.16 &  283.63 & 2.99 &  9882.98 &  549.05 & 0.09 \\
\midrule
\multirow{4}{*}{\textbf{Llama}} 
 & \multirow{2}{*}{Python} & mean 
    & 5.80 & 17.89 & 15.24 & 29.68 & 23.69 & 44.93 &  95.89 & 222.09 & 4.78 & 1582.12 & 87.90 & 0.07 \\
 &                         & std  
    & 1.27 &  6.22 &  6.22 & 12.21 &  7.12 & 18.42 &  41.73 & 112.55 & 1.44 & 1201.02 & 66.72 & 0.04 \\
 \cmidrule(lr){2-15}
 & \multirow{2}{*}{C/C++}  & mean 
    & 7.76 & 47.56 & 71.16 & 244.13 & 55.31 & 315.29 & 291.68 & 1869.23 & 19.49 & 52663.32 & 2925.74 & 0.62 \\
 &                         & std  
    & 1.09 &  7.90 & 21.61 &  70.48 &  8.37 &  88.71 &  60.37 &  579.59 & 5.68 & 49067.76 & 2725.99 & 0.19 \\
\midrule
\multirow{4}{*}{\textbf{Qw4B}} 
 & \multirow{2}{*}{Python} & mean 
    & 5.20 & 15.25 & 12.87 & 24.97 & 20.45 & 37.84 &  78.39 & 178.54 & 4.27 & 1105.91 & 61.44 & 0.06 \\
 &                         & std  
    & 0.85 &  4.14 &  3.87 &  7.66 &  4.77 & 11.52 &  28.29 &  70.75 & 0.91 &  651.17 & 36.18 & 0.02 \\
 \cmidrule(lr){2-15}
 & \multirow{2}{*}{C/C++}  & mean 
    & 7.47 & 50.81 & 66.06 & 235.00 & 58.28 & 301.07 & 315.46 & 1822.14 & 16.76 & 41082.45 & 2282.36 & 0.61 \\
 &                         & std  
    & 0.64 &  7.44 & 12.31 &  48.69 &  7.68 & 59.78 & 58.12 & 424.64 & 2.86 & 18516.05 & 1028.67 & 0.14 \\
\midrule
\multirow{4}{*}{\textbf{Qw8B}} 
 & \multirow{2}{*}{Python} & mean 
    & 5.11 & 15.05 & 12.86 & 24.96 & 20.16 & 37.82 &  77.21 & 180.06 & 4.19 & 1174.21 & 65.23 & 0.06 \\
 &                         & std  
    & 1.09 &  4.86 &  4.92 &  9.69 &  5.65 & 14.60 &  32.15 &  86.64 & 1.21 &  852.52 & 47.36 & 0.03 \\
 \cmidrule(lr){2-15}
 & \multirow{2}{*}{C/C++}  & mean 
    & 7.29 & 47.40 & 64.41 & 223.15 & 54.69 & 287.56 & 289.40 & 1719.90 & 16.42 & 42327.90 & 2351.55 & 0.57 \\
 &                         & std  
    & 0.91 &  7.83 & 18.62 &  71.91 &  8.26 & 89.09 & 60.75 & 643.36 & 3.90 & 43502.31 & 2416.80 & 0.21 \\
\midrule
\multirow{4}{*}{\textbf{Qw14B}} 
 & \multirow{2}{*}{Python} & mean 
    & 5.64 & 16.86 & 14.27 & 27.67 & 22.50 & 41.93 &  88.48 & 201.97 & 4.62 & 1330.55 &  73.92 & 0.07 \\
 &                         & std  
    & 0.90 &  4.24 &  4.16 &  8.17 &  4.90 & 12.32 &  28.49 &  74.61 & 0.99 &  763.02 &  42.39 & 0.02 \\
 \cmidrule(lr){2-15}
 & \multirow{2}{*}{C/C++}  & mean 
    & 7.78 & 55.85 & 72.02 & 255.75 & 63.62 & 327.77 & 354.15 & 2021.52 & 17.46 & 46444.34 & 2580.24 & 0.67 \\
 &                         & std  
    & 0.73 &  8.96 & 14.89 &  53.41 &  9.29 &  67.10 &  70.76 &  482.39 & 3.33 & 23781.57 & 1321.20 & 0.16 \\
\midrule
\multirow{4}{*}{\textbf{Qw32B}} 
 & \multirow{2}{*}{Python} & mean 
    & 5.90 & 17.56 & 14.72 & 28.50 & 23.46 & 43.22 &  93.84 & 211.90 & 4.76 & 1458.62 & 81.03 & 0.07 \\
 &                         & std  
    & 1.14 &  5.01 &  4.87 &  9.50 &  5.83 & 14.35 &  33.83 &  87.12 & 1.22 &  921.81 & 51.21 & 0.03 \\
 \cmidrule(lr){2-15}
 & \multirow{2}{*}{C/C++}  & mean 
    & 8.04 & 54.25 & 72.48 & 260.03 & 62.29 & 332.51 & 342.14 & 2038.57 & 18.82 & 51082.68 & 2837.93 & 0.68 \\
 &                         & std  
    & 0.89 &  8.35 & 15.50 &  56.92 &  8.73 & 70.74 & 64.18 & 488.85 & 4.02 & 23772.82 & 1320.71 & 0.16 \\
\midrule

\multirow{4}{*}{\textbf{QwCo}} 
 & \multirow{2}{*}{Python} & mean 
    & 5.51 & 17.09 & 14.74 & 28.57 & 22.60 & 43.31 &  90.83 & 212.60 & 4.58 & 1457.38 & 80.97 & 0.07 \\
 &                         & std  
    & 1.27 &  6.30 &  6.22 & 12.17 &  7.22 & 18.37 &  42.43 & 112.54 & 1.45 & 1142.83 & 63.49 & 0.04 \\
 \cmidrule(lr){2-15}
 & \multirow{2}{*}{C/C++}  & mean 
    & 7.70 & 43.79 & 64.09 & 216.48 & 51.49 & 280.57 & 264.30 & 1633.43 & 18.53 & 41956.16 & 2330.90 & 0.54 \\
 &                         & std  
    & 1.04 &  4.85 & 16.98 &  57.39 &  5.31 & 72.14 & 36.29 & 455.12 & 5.03 & 27470.03 & 1526.11 & 0.15 \\
    \midrule
\multirow{4}{*}{\textbf{Human}} 
  & \multirow{2}{*}{Python} & mean  
      & 5.89  & 20.20  & 17.30  & 33.84  & 26.09  
      & 51.14  & 111.90  & 263.59  & 4.96  
      & 1973.81  & 109.66  & 0.09 \\
  &                          & std   
      & 0.83  &  4.00  &  3.72  &  7.41  &  4.58  
      & 11.13  &  25.66  &  66.45  & 0.85  
      &  624.38  &  34.69  & 0.02 \\
  \cmidrule(lr){2-15}
  & \multirow{2}{*}{C/C++}   & mean  
      & 9.66  & 67.54   & 116.54 & 387.92 & 77.20  
      & 504.47 & 454.55  & 3303.25 & 26.78 
      & 145862.09 & 8103.45 & 1.10 \\
  &                          & std   
      & 1.69  & 31.47  &  98.16 & 254.47 & 32.55  
      & 340.85 & 259.09  & 2649.44 & 10.24 
      & 235927.01 & 13107.06 & 0.88 \\

\bottomrule
\end{tabular}

\end{adjustbox}

\caption{Halstead metrics on the evaluation subset. Model abbreviations: Claude (\textit{Claude-3.5-Sonnet}), DSV3 (\textit{DeepSeek-V3}), DSR1 (\textit{DeepSeek-R1}), Gemma (\textit{Gemma-3-27B}), GPT (\textit{GPT-4o-mini}), Llama (\textit{Llama-3.3-Nemotron-Super-49B-v1}), QwxB (\textit{Qwen3-xB}), and QwCo (\textit{Qwen2.5-Coder-32B-Instruct}). The label \textit{Human} refers to human-written code. Metric abbreviations: Cal\_Len (calculated program length), Difficul (\textit{difficulty}), Time\_Sec (time to implement), and Bugs (estimated bugs).}
\label{tab:halstead_status_subset}
\end{table*}

\begin{table*}[!t]
\small
\renewcommand{\arraystretch}{1.1}
\setlength{\tabcolsep}{4pt}
\centering
\begin{tabular}{l l l|ccccccc}
\toprule
\textbf{Model} & \textbf{Language} & \textbf{Stat.} 
  & \textbf{Volume} & \textbf{Cyclomatic} & \textbf{SLOC} 
  & \textbf{LLOC} & \textbf{Comment\_Rate} & \textbf{MI\_Std} & \textbf{MI\_Custom} \\
\midrule
\multirow{4}{*}{\textbf{Claude}} 
  & \multirow{2}{*}{Python} & mean 
      & 277.88  & 9.37  & 29.06 & 29.22 & 20.47 & 72.62 & 72.72 \\
  &                          & std  
      & 164.09  & 4.71  &  9.71 &  9.70 & 11.13 &  7.45 &  7.45 \\
  \cmidrule(lr){2-10}
  & \multirow{2}{*}{C/C++}   & mean 
      & 2611.66 & 11.83 & 49.28 & 40.61 &  0.09 & 53.70 & 39.47 \\
  &                          & std  
      & 698.95  &  3.55 & 10.44 &  9.13 &  0.05 &  5.63 &  2.71 \\
\midrule
\multirow{4}{*}{\textbf{DSV3}} 
  & \multirow{2}{*}{Python} & mean 
      & 254.75  & 4.16  & 27.69 & 27.88 &  8.96 & 63.28 & 63.95 \\
  &                         & std  
      & 85.88   & 1.91  &  5.54 &  5.59 &  8.35 &  4.72 &  4.68 \\
  \cmidrule(lr){2-10}
  & \multirow{2}{*}{C/C++}   & mean 
      & 1908.68 & 10.37 & 42.54 & 37.72 &  0.00 & 42.87 & 42.02 \\
  &                         & std  
      & 305.03  & 2.10  &  5.73 &  5.55 &  0.01 &  2.06 &  1.59 \\
\midrule
\multirow{4}{*}{\textbf{DSR1}} 
  & \multirow{2}{*}{Python} & mean 
      & 319.45  & 5.63  & 33.16 & 33.62 & 18.92 & 56.58 & 57.29 \\
  &                         & std  
      & 147.19  & 3.53  &  9.56 &  9.67 & 47.74 &  7.55 &  7.49 \\
  \cmidrule(lr){2-10}
  & \multirow{2}{*}{C/C++}   & mean 
      & 2720.12 & 12.13 & 48.28 & 41.80 &  0.02 & 42.08 & 40.41 \\
  &                         & std  
      & 1753.39 &  3.52 & 12.44 &  9.98 &  0.05 &  3.89 &  2.91 \\
\midrule
\multirow{4}{*}{\textbf{Gemma}} 
  & \multirow{2}{*}{Python} & mean 
      & 278.83  & 11.64 & 31.14 & 31.49 &  0.54 & 52.74 & 52.74 \\
  &                         & std  
      & 117.70  &  3.76 &  6.82 &  6.89 &  0.63 &  3.80 &  3.79 \\
  \cmidrule(lr){2-10}
  & \multirow{2}{*}{C/C++}   & mean 
      & 2379.33 & 18.64 & 51.23 & 47.04 &  0.00 & 38.88 & 38.23 \\
  &                         & std  
      & 474.05  &  5.14 &  8.57 &  8.60 &  0.00 &  2.89 &  2.51 \\
\midrule
\multirow{4}{*}{\textbf{GPT}} 
  & \multirow{2}{*}{Python} & mean 
      & 151.65  &  4.34 & 19.77 & 20.18 &  7.35 & 66.81 & 67.11 \\
  &                         & std  
      & 59.85   &  2.38 &  3.89 &  3.99 &  8.25 &  8.40 &  8.40 \\
  \cmidrule(lr){2-10}
  & \multirow{2}{*}{C/C++}   & mean 
      & 1571.22 &  8.01 & 33.61 & 28.85 &  0.02 & 48.89 & 44.51 \\
  &                         & std  
      & 283.63  &  1.75 &  4.63 &  4.22 &  0.03 &  5.41 &  1.87 \\
\midrule
\multirow{4}{*}{\textbf{Llama}} 
  & \multirow{2}{*}{Python} & mean 
      & 222.09  &  6.67 & 24.25 & 24.67 &  3.60 & 59.23 & 59.45 \\
  &                         & std  
      & 112.55  &  3.85 &  7.10 &  7.27 &  7.10 &  7.05 &  7.04 \\
  \cmidrule(lr){2-10}
  & \multirow{2}{*}{C/C++}   & mean 
      & 1869.23 & 10.02 & 39.97 & 34.97 &  0.00 & 43.38 & 42.66 \\
  &                         & std  
      & 579.59  &  3.80 &  9.92 &  9.21 &  0.01 &  3.50 &  3.05 \\
\midrule
\multirow{4}{*}{\textbf{Qw4B}} 
  & \multirow{2}{*}{Python} & mean 
      & 178.54  &  5.43 & 25.25 & 25.61 & 14.78 & 66.60 & 66.90 \\
  &                         & std  
      & 70.75   &  1.80 &  4.61 &  4.63 & 11.42 &  5.08 &  5.07 \\
  \cmidrule(lr){2-10}
  & \multirow{2}{*}{C/C++}   & mean 
      & 1822.14 &  9.85 & 38.15 & 33.42 &  0.02 & 46.83 & 43.38 \\
  &                         & std  
      & 424.64  &  2.20 &  6.74 &  5.96 &  0.02 &  3.38 &  2.02 \\
\midrule
\multirow{4}{*}{\textbf{Qw8B}} 
  & \multirow{2}{*}{Python} & mean 
      & 180.06  &  4.63 & 23.17 & 23.57 &  6.59 & 62.92 & 63.26 \\
  &                         & std  
      & 86.64   &  2.23 &  5.84 &  5.91 &  8.36 &  6.05 &  6.03 \\
  \cmidrule(lr){2-10}
  & \multirow{2}{*}{C/C++}   & mean 
      & 1719.90 &  8.72 & 36.59 & 31.88 &  0.02 & 46.74 & 44.08 \\
  &                         & std  
      & 643.36  &  3.47 &  9.41 &  8.86 &  0.03 &  3.89 &  2.63 \\
\midrule
\multirow{4}{*}{\textbf{Qw14B}} 
  & \multirow{2}{*}{Python} & mean 
      & 201.97  &  5.68 & 27.68 & 28.20 &  9.60 & 64.48 & 64.77 \\
  &                         & std  
      & 74.61   &  1.97 &  5.15 &  5.20 &  7.13 &  4.72 &  4.72 \\
  \cmidrule(lr){2-10}
  & \multirow{2}{*}{C/C++}   & mean 
      & 2021.52 &  9.74 & 41.83 & 36.14 &  0.03 & 47.95 & 42.10 \\
  &                         & std  
      & 482.39  &  2.37 &  7.65 &  6.87 &  0.03 &  3.94 &  2.15 \\
\midrule

\multirow{4}{*}{\textbf{Qw32B}} 
  & \multirow{2}{*}{Python} & mean 
      & 211.90  &  4.71 & 26.86 & 27.27 &  4.78 & 61.56 & 62.00 \\
  &                         & std  
      & 87.12   &  2.45 &  6.03 &  6.09 &  4.69 &  5.79 &  5.76 \\
  \cmidrule(lr){2-10}
  & \multirow{2}{*}{C/C++}   & mean 
      & 2038.57 & 10.26 & 41.78 & 37.35 &  0.02 & 45.98 & 42.01 \\
  &                         & std  
      & 488.85  &  2.73 &  8.41 &  8.02 &  0.02 &  4.43 &  2.52 \\
\midrule

\multirow{4}{*}{\textbf{QwCo}} 
  & \multirow{2}{*}{Python} & mean 
      & 212.60  &  6.19 & 23.45 & 23.93 &  1.26 & 57.81 & 57.94 \\
  &                         & std  
      & 112.54  &  3.00 &  7.00 &  7.13 &  3.01 &  5.88 &  5.90 \\
  \cmidrule(lr){2-10}
  & \multirow{2}{*}{C/C++}   & mean 
      & 1633.43 &  9.22 & 34.53 & 30.81 &  0.00 & 44.76 & 44.44 \\
  &                         & std  
      & 455.12  &  2.97 &  8.02 &  7.71 &  0.00 &  3.13 &  2.84 \\
\midrule
\multirow{4}{*}{\textbf{Human}} 
  & \multirow{2}{*}{Python} & mean 
      & 263.59  &  2.64 & 36.01 & 23.59 &  6.51 & 60.82 & 60.82 \\
  &                         & std  
      & 66.45   &  1.45 &  9.26 &  4.76 & 11.64 &  5.64 &  5.64 \\
  \cmidrule(lr){2-10}
  & \multirow{2}{*}{C/C++}   & mean 
      & 3303.25 & 12.62 & 55.01 & 39.09 &  0.05 & 45.10 & 38.82 \\
  &                         & std  
      & 2649.44 &  7.00 & 29.40 & 18.13 &  0.09 &  7.81 &  6.46 \\
\bottomrule
\end{tabular}
\caption{Maintainability results on the evaluation subset, broken down by model, language (Python vs.\ C/C++), and statistic (mean vs.\ std). Model abbreviations follow those in Table~\ref{tab:halstead_status_subset}.}
\label{tab:maintainability_status_subset}
\end{table*}

\end{document}